\documentclass{ceurart}

\usepackage{amsmath}
\usepackage{amsthm}
\usepackage{amssymb}
\usepackage{booktabs}
\usepackage{algorithm}
\usepackage{algorithmic}
\usepackage{float}
\usepackage{xcolor}

\definecolor{diffcolor}{rgb}{0.16, 0.32, 0.75}

\begin{document}

\copyrightyear{2026}
\copyrightclause{Copyright for this paper by its authors. Use permitted under Creative Commons License Attribution 4.0 International (CC BY 4.0).}
\conference{EXPLIMED 2026 - Third Workshop on Explainable Artificial Intelligence for the medical domain - 15-17 August 2026, Bremen, Germany}

\title{Explaining BiomedCLIP with Weighted Banzhaf Interactions Supported by Tree-Gram Parsing}

\author[1,2]{Jakub Rymarski}[email=jr472844@students.mimuw.edu.pl]
\author[1]{Adam Rempała}[email=ar406309@students.mimuw.edu.pl]
\author[1,2]{Bartłomiej Sobieski}[email=sobieski.bartlomiej.jan@gmail.com]
\author[1,2]{Przemysław Biecek}[email=przemyslaw.biecek@gmail.com]

\address[1]{University of Warsaw, Poland}
\address[2]{Centre for Credible AI, Warsaw University of Technology, Poland}

\begin{abstract}
    Vision-Language Models (VLMs) are demonstrating significant capabilities in medical tasks like radiology analysis, yet providing faithful and interpretable explanations remains a key consideration for their responsible deployment in clinical settings. However, existing explanation methods, such as the widely used FIxLIP framework, often struggle with the fine-grained nature of modern tokenizers. The tokenization problem fragments clinical concepts—splitting terms like ``saddle embolus'' into scattered, meaningless subwords—which leads to noisy, semantically incoherent cross-modal attributions. Such fragmentation also results in a combinatorial explosion of interaction possibilities, obscuring the model's true reasoning. To address this, we introduce ParseFIxLIP, an extension that incorporates the Tree-Gram Parsing into the Banzhaf interaction game used by FIxLIP. This semantically informed strategy utilizes dependency parsing trees to define explanation players by grouping related text tokens into semantically coherent units. Our \texttt{smart\_depth} grouping strategy, merging tokens according to spaCy token dependency tree, successfully mitigates concept fragmentation, yielding substantially more interpretable cross-modal interactions by unifying complex medical concepts. Quantitatively, while baselines struggled with the high dimensionality of long captions, our parsing approach maintained statistical robustness and semantic parsimony. Qualitative analysis on BiomedCLIP, validated on medical imagery (ROCOv2) and general examples, confirms that the approach accurately captures the synergistic influence of grouped words on model predictions. In conclusion, our work offers intuitive and clinically relevant insights into VLM decision-making, fulfilling the critical need for coherent explanations in the medical domain.
\end{abstract}


\maketitle

\section{Introduction}

Contrastive language–image pre-training has revolutionized the development of powerful Vision–Language Models (VLMs), enabling them to perform exceptionally well on a wide range of downstream tasks~\cite{radford2021learning}. Examples include setting new benchmarks in zero-shot classification~\cite{jia2021scaling} and complex visual question answering~\cite{li2022blip}. In the medical domain, models such as BiomedCLIP~\cite{zhang2023biomedclip} demonstrate strong zero-shot capabilities in processing complex radiological image–text pairs. However, these encoders remain highly opaque. Deploying them in high-stakes clinical decision-making requires not only high accuracy but also trustworthy and interpretable explanations of how predictions arise from specific visual and textual features~\cite{acosta2022multimodal, amann2020explainability}.

One way to validate these similarity predictions is through game-theoretic explainability frameworks that build upon established research on interaction indices and generalized additive models~\cite{bordt2023shapley, fumagalli2023shapiq, pelegrina2023additive}, such as the recently proposed FIxLIP~\cite{baniecki2025fixlip}. By modeling input tokens as players in a cooperative game, FIxLIP faithfully decomposes the model's output into main effects and pairwise cross-modal interactions using Weighted Banzhaf values~\cite{marichal2011weighted}.  Yet, a fundamental limitation of applying such methods directly to modern VLMs lies in their reliance on subword tokenizers. In a clinical context, processing raw tokens leads to severe concept fragmentation—splitting precise medical terminology like ``saddle embolus'' into scattered, arbitrary fragments (e.g., ``sad'', ``dle'', ``embol'', ``us''). This granularity creates two critical problems: it generates semantically incoherent explanations that are unreadable for clinicians, and it triggers a combinatorial explosion of meaningless cross-modal interactions. Consequently, the explanation process suffers from multicollinearity and the curse of dimensionality, heavily penalizing goodness-of-fit metrics~\cite{covert2021improving, hastie2009elements}, especially when analyzing long, complex radiology reports.

To overcome these challenges, we introduce \textbf{ParseFIxLIP}, a model-agnostic approach that incorporates linguistic structure into interaction-based explanations. Building upon the concept of Tree-Gram Parsing~\cite{simaan2000tree}, we utilize Natural Language Processing (NLP) dependency parsing (e.g., via spaCy~\cite{spacy}) to formally project the original Banzhaf game into a lower-dimensional quotient game. By linking syntactically dependent modifiers to their roots, we redefine the players of the game as semantically coherent textual units. This enforces semantic parsimony, structurally eliminating the noise caused by subword fragmentation before the game-theoretic interactions are even computed. 

\textbf{Contributions.} Our work advances the state of VLM interpretability in the following ways: \textbf{(1) A linguistically-informed game-theoretic explanation.} We introduce ParseFIxLIP, adapting the Banzhaf interaction framework by using dependency parsing trees. To our knowledge, this is the first work to address the token fragmentation problem in game-theoretic XAI by grouping tokens into coherent clinical concepts. \textbf{(2) Theoretical mitigation of the dimensionality curse.} We demonstrate that our quotient game formulation inherently solves the multicollinearity and high-dimensionality issues present in standard token-level interaction methods. By reducing the number of players, we stabilize the Weighted Banzhaf estimation under a fixed sampling budget. \textbf{(3) Empirical validation on medical data.} We evaluate multiple semantic grouping strategies on the ROCOv2 dataset~\cite{ruckert2024rocov2}. Quantitative metrics show that while standard baselines fail (yielding negative Adjusted $R^2$) on long radiology captions, our semantic grouping maintains statistical robustness.

\section{Related Work}

\textbf{Vision-Language Models in the Medical Domain.} The success of contrastive language-image pre-training~\cite{radford2021learning} has led to its rapid adoption in specialized domains, notably medicine~\cite{acosta2022multimodal, moor2023foundation}. Models such as BiomedCLIP~\cite{zhang2023biomedclip} leverage massive datasets of scientific image-text pairs to achieve strong zero-shot performance in tasks like radiology report analysis. However, the black-box nature of these multimodal architectures poses a significant barrier to clinical adoption. While traditional saliency maps have been applied to medical imaging to validate model predictions, they often fail to capture the nuanced, cross-modal dependencies critical to clinical reasoning, driving the need for more sophisticated, multimodal explainability frameworks.

\textbf{Game-Theoretic Explainability for VLMs.} To provide faithful feature attribution, game-theoretic methods rooted in Shapley and Banzhaf values have become increasingly prominent in machine learning~\cite{covert2021explaining}. In the multimodal space, existing gradient and attention based methods~\cite{chefer2021generic, zhao2024grad} generally approximate only first-order attributions, acting as simple unimodal saliency maps. To capture synergistic cross-modal effects, the FIxLIP framework~\cite{baniecki2025fixlip} introduced a computationally efficient approximation of Weighted Banzhaf Interactions for vision--language encoders. While FIxLIP successfully uncovers pairwise interactions between image patches and text, its direct application to medical models is severely limited by its reliance on raw subword tokens, which struggle to represent complex clinical terminology.

\textbf{Linguistic Structure and Concept-Based Interpretability.} The reliance of modern VLMs on subword tokenizers (e.g., BPE, WordPiece) creates a well-documented disconnect between mathematical model inputs and human-understandable concepts~\cite{mielke2021between, bostrom2020byte}. In NLP, techniques like Tree-Gram Parsing~\cite{simaan2000tree} and dependency parsing have long been utilized to group words into modular semantic units. However, their integration into multimodal explainability remains largely unexplored. Previous attempts at concept-level explanations typically require auxiliary models, human-annotated concepts, or extensive linear probing~\cite{yuksekgonul2022post, oikarinen2022clip}. Our approach, ParseFIxLIP, bridges this gap by directly embedding NLP dependency trees into the game-theoretic formulation.

\section{Methodology}
\label{sec:methodology}

\subsection{Background: Vision-Language Similarity and the FIxLIP Game}
A vision--language encoder $f$ consists of a vision encoder $f_I : \mathbb{R}^{n_I} \to \mathbb{R}^d$ and a language encoder $f_T : \mathbb{R}^{n_T} \to \mathbb{R}^d$. For a given input pair of $n_I$ image patches $\mathbf{x}_I \in \mathbb{R}^{n_I}$ and $n_T$ text tokens $\mathbf{x}_T \in \mathbb{R}^{n_T}$, the model computes the cosine similarity $f(\mathbf{x}_I, \mathbf{x}_T)$ of their $d$-dimensional embeddings. Let $N = N_I \cup N_T$ denote the combined index set of raw input features, where $N_I = \{1, \dots, n_I\}$ represents the image patches and $N_T = \{n_I+1, \dots, n_I+n_T\}$ represents the text subword tokens produced by a standard tokenizer.

The FIxLIP framework explains this similarity prediction by formulating it as a cooperative game. For a given input pair $(\mathbf{x}_I, \mathbf{x}_T)$ and a baseline $(\mathbf{b}_I, \mathbf{b}_T)$, the masking operator $\oplus_S$ retains the features in subset $S$ and replaces the hidden rest with their baseline values. For any subset of visible tokens (a mask) $M \subseteq N$, the model yields a similarity score:
\begin{equation}
    \nu(M) = f(\mathbf{x}_I \oplus_{M \cap N_I} \mathbf{b}_I, \mathbf{x}_T \oplus_{M \cap N_T} \mathbf{b}_T)
\end{equation}
FIxLIP decomposes this score into main effects and pairwise interactions by fitting a 2-additive game approximation $\hat{\nu}_{\mathbf{e}}(M)$:
\begin{equation}
    \hat{\nu}_{\mathbf{e}}(M) = e_0 + \sum_{i \in M} e_i + \sum_{\{i,j\} \subseteq M: i \neq j} e_{\{i,j\}}
\end{equation}
where $e_0$ is a constant baseline, $e_i$ is the individual attribution of token $i$, and $e_{\{i,j\}}$ represents the pairwise interaction between tokens $i$ and $j$. The explanation parameters are found by minimizing a $p$-weighted least squares error (approximating Weighted Banzhaf Interactions) over uniformly sampled masks. Explicitly, given a computational budget of $K$ masks sampled uniformly at random, $M_1, \dots, M_K \sim \mathcal{U}(2^N)$, we minimize the empirical loss:
\begin{equation}
    \hat{\mathcal{L}}(\mathbf{e}) = \frac{1}{K} \sum_{k=1}^K p(M_k) \left( \nu(M_k) - \hat{\nu}_{\mathbf{e}}(M_k) \right)^2
\end{equation}
where the weighting term $p(M_k) = p^{|M_k|}(1-p)^{|N|-|M_k|}$ reflects the chosen probability $p \in (0, 1)$ of an individual token remaining visible.

\subsection{The Fragmentation Problem and Multicollinearity}
While FIxLIP provides a rigorous explanation basis, applying it directly to VLMs at the subword level introduces critical statistical and semantic flaws. Modern tokenizers split complex clinical terms into arbitrary fragments (e.g., ``saddle'' into ``sad'' and ``dle''). 

In the random masking process, evaluating a state where ``sad'' is visible but ``dle'' is hidden creates an out-of-distribution and meaningless text input. Furthermore, from a theoretical perspective, artificially splitting a single underlying concept into multiple sub-tokens creates highly correlated variables in the interaction space. This multicollinearity destabilizes the Banzhaf regression estimation. As the length of the medical report increases, the dimensionality of the raw token set $|N|$ grows. Driven by the curse of dimensionality, the combinatorial space of possible masks explodes. Under a fixed computational budget of sampled masks, the regression fails to isolate meaningful signals from the noise, resulting in poor goodness-of-fit.

\subsection{ParseFIxLIP: Tree-Gram Parsing and Semantic Grouping}
To solve both the semantic fragmentation and the mathematical instability, we introduce ParseFIxLIP. Instead of computing interactions on the raw token set $N$, we project the explanation task into a reduced dimensional space. We define a surjective mapping function $\pi: N \rightarrow N'$ that groups raw text tokens into a smaller set of semantically coherent macro-players $N'$, based on the Tree-Gram Parsing paradigm~\cite{simaan2000tree}.

This transformation defines a \textit{reduced explanation game} $\mathcal{G}' = (N', \nu')$. For any subset of macro-players $M' \subseteq N'$, the corresponding similarity score is evaluated by mapping the macro-mask back to the raw tokens:
\begin{equation}
    \nu'(M') = \nu(\{i \in N \mid \pi(i) \in M'\})
\end{equation}
When a macro-player in $N'$ is masked, all its constituent sub-tokens from $N$ are masked simultaneously. To construct the mapping $\pi$, we utilize the spaCy dependency parser to implement four grouping strategies: \textbf{(1) \texttt{raw}:} The baseline tokenizer splitting (identity mapping). \textbf{(2) \texttt{words}:} Subwords are strictly merged into full dictionary words. \textbf{(3) \texttt{noun\_chunks}:} Multi-word noun phrases are grouped into single entities, leaving remaining tokens as standalone players. \textbf{(4) \texttt{smart\_depth}:} Our novel syntactic strategy, where heads of dependency trees are recursively grouped with their modifiers.

\textbf{Theoretical justification.} Formulating the explanation on a reduced game space provides a strong theoretical guarantee for long reports. By enforcing semantic grouping, we ensure that $|N'| \ll |N|$. Drastically reducing the number of players under a fixed sampling budget (e.g., $K=50,000$ perturbations) directly reduces the variance of weighted least squares estimator. This effectively neutralizes the multicollinearity problem and maintains semantic parsimony, structurally guaranteeing a higher and more stable Adjusted $R^2$ fit compared to raw token baselines.

\section{Quantitative Experiments}

We evaluate the proposed parsing strategies on a subset of the ROCOv2 dataset following
the evaluation protocol of~\cite{baniecki2025fixlip}. Since explanation quality strongly depends on report length, we stratify the analysis by caption length.

The faithfulness is measured by Spearman’s Rank Correlation and Adjusted $R^2$ to check if Tree-Gram Parsing increases it. Spearman’s correlation assesses the general capability of the explanation method to recover the ranking of similarity scores for uniformly masked inputs compared to the original model predictions. Adjusted $R^2$ is used to evaluate the precise goodness-of-fit. It accounts for the varying number of players resulting from different parsing strategies. 

\begin{figure}[htbp]
    \centering
    \begin{minipage}{0.49\linewidth}
        \centering
        \includegraphics[width=\linewidth]{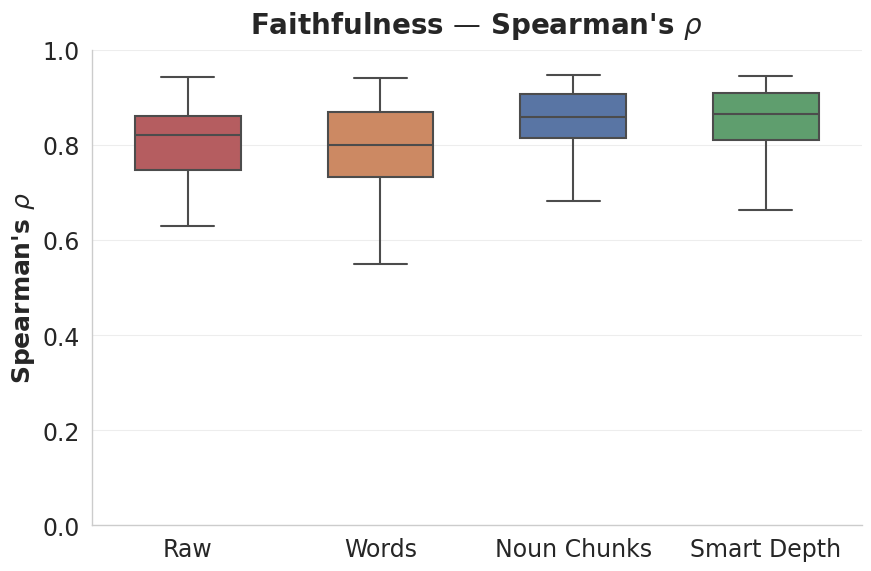} 
    \end{minipage}
    \hfill
    \begin{minipage}{0.49\linewidth}
        \centering
        \includegraphics[width=\linewidth]{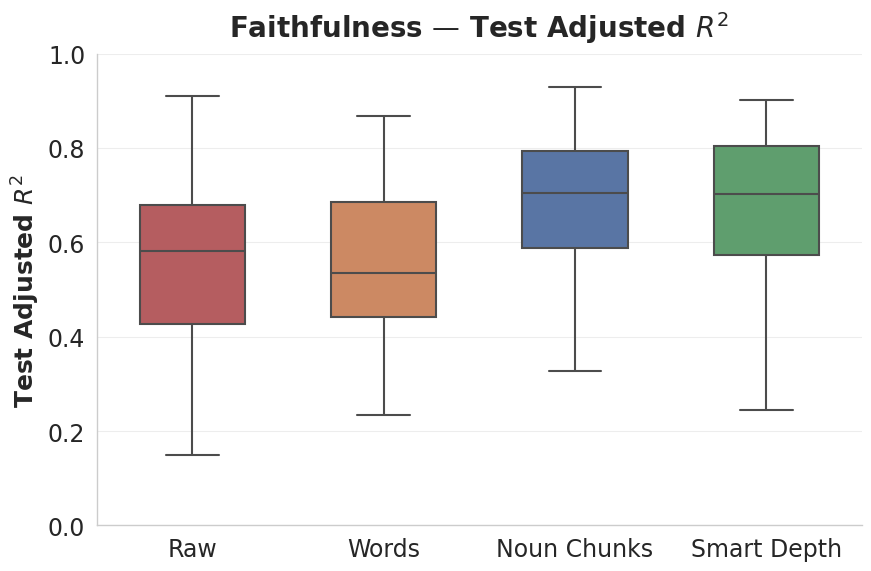} 
    \end{minipage}
    \caption{Faithfulness evaluation on 128 short captions samples. \textbf{Left:} Spearman Rank Correlation. \textbf{Right:} Adjusted $R^2$. \texttt{noun\_chunks} and \texttt{smart\_depth} \textbf{consistently outperform the token-level baselines}.}
    \label{fig:faithfulness_short}
\end{figure}

In terms of Spearman correlation on short captions, \texttt{noun\_chunks} and \texttt{smart\_depth} demonstrate the highest stability with a median score close to 0.9. This disparity is more visible in the Adjusted $R^2$ metric, where both \texttt{smart\_depth} and \texttt{noun\_chunks} achieve median values above 0.7 (Figure \ref{fig:faithfulness_short}).

We extend our quantitative evaluation to the challenging subset of long captions ($> 30$ words). This setting results in a high-dimensional feature space populated largely by functional words and background noise. The interaction space also expands drastically. We employed a sampling budget of $N = 50,000$ perturbations. The following results suggest that this budget might be insufficient for the token-level baseline to stabilize the signal-to-noise ratio in such a sparse feature space.

\begin{figure}[htbp]
    \centering
    \begin{minipage}{0.49\linewidth}
        \centering
        \includegraphics[width=\linewidth]{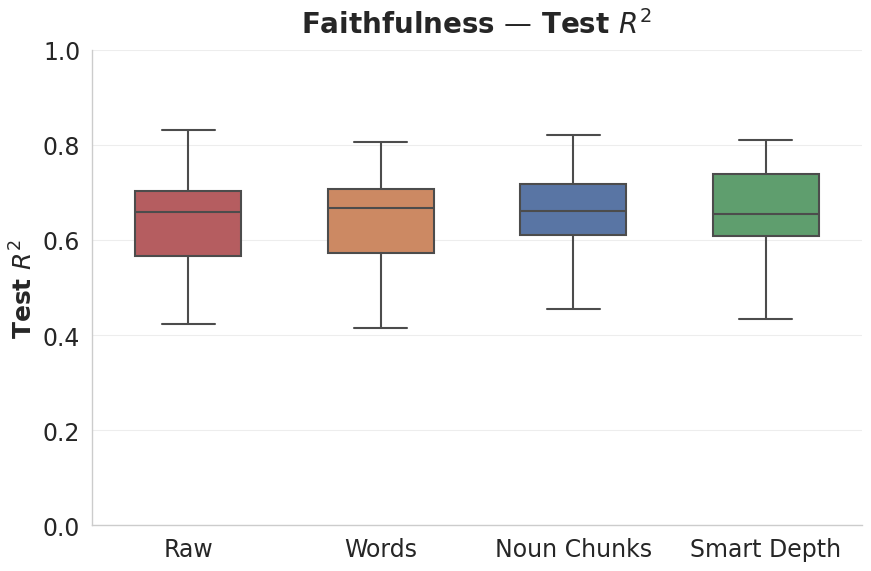}
    \end{minipage}
    \hfill
    \begin{minipage}{0.49\linewidth}
        \centering
        \includegraphics[width=\linewidth]{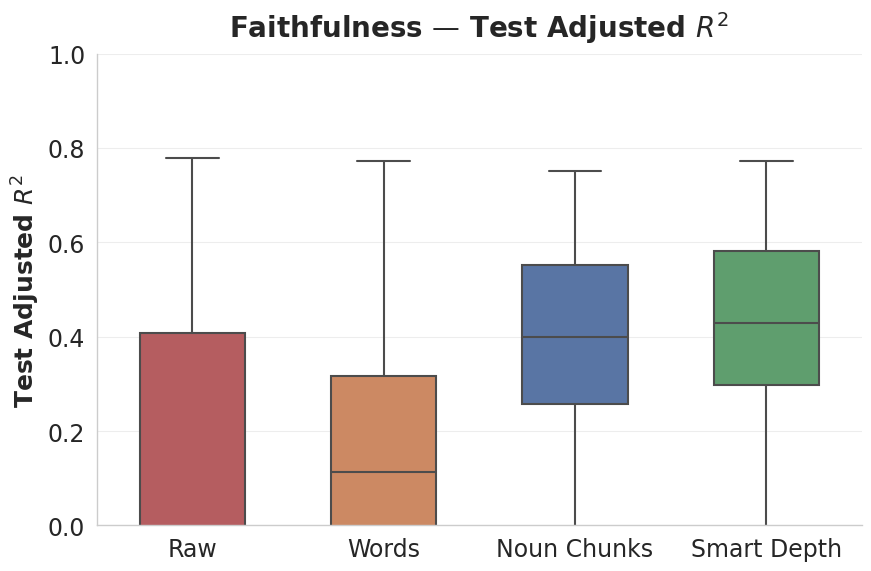}
    \end{minipage}
    \caption{Goodness-of-fit metrics on long captions. \textbf{Left:} Standard $R^2$ shows comparable explanatory potential. \textbf{Right:} Adjusted $R^2$ reveals the inefficiency of the \texttt{raw} strategy, which is penalized for its high feature count, whereas \texttt{smart\_depth} remains robust due to semantic parsimony.}
    \label{fig:long_captions_r2}
\end{figure}

First, we analyze the model fit in Figure \ref{fig:long_captions_r2}. Evaluated using the Friedman test and subsequent pairwise Wilcoxon signed-rank tests at a significance level of $\alpha = 0.05$, the Standard $R^2$ (Left) shows statistically significant differences across strategies ($p < 0.001$). However, a critical divergence appears in the Adjusted $R^2$ (Right), where semantic grouping drastically outperforms the raw token baseline ($p < 0.0001$). The \texttt{raw} strategy drops to negative values for two samples. This phenomenon likely reflects the curse of dimensionality, as the \texttt{raw} method attempts to fit a linear explanation using over 70 variables (tokens + patches). The penalty for this high complexity, combined with the multicollinearity of noisy linguistic tokens, outweighs the marginal gains in fit. In contrast, \texttt{smart\_depth} effectively performs dimensionality reduction, condensing the input into $< 15$ semantic concepts. This allows the method to recover a statistically robust signal.

We further assess the quality of the explanations, to check if Tree-Gram Parsing improves them, using Insertion and Deletion curves, quantified by the Area Between Curves
(AID). These metrics measure how rapidly the model’s confidence changes as important features are added to or removed from the input. 

\begin{figure}[htbp]
    \centering
    \begin{minipage}{0.49\linewidth}
        \centering
        \includegraphics[width=\linewidth]{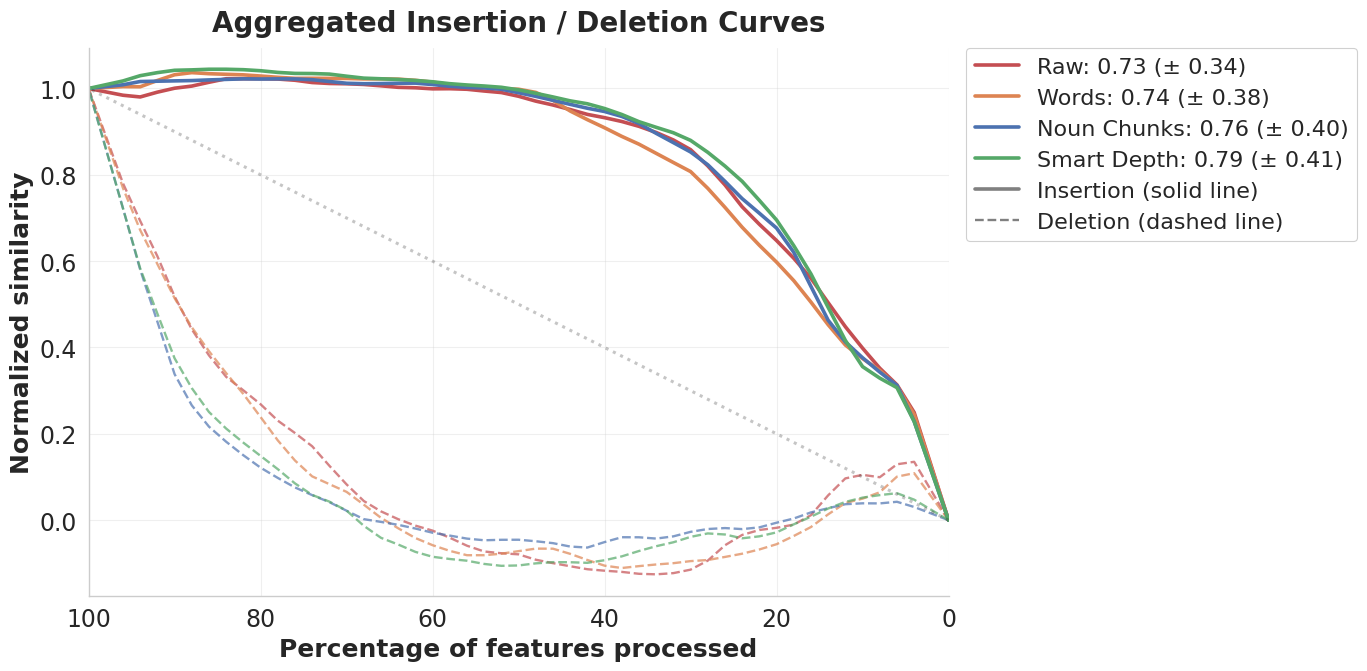}
    \end{minipage}
    \hfill
    \begin{minipage}{0.49\linewidth}
        \centering
        \includegraphics[width=\linewidth]{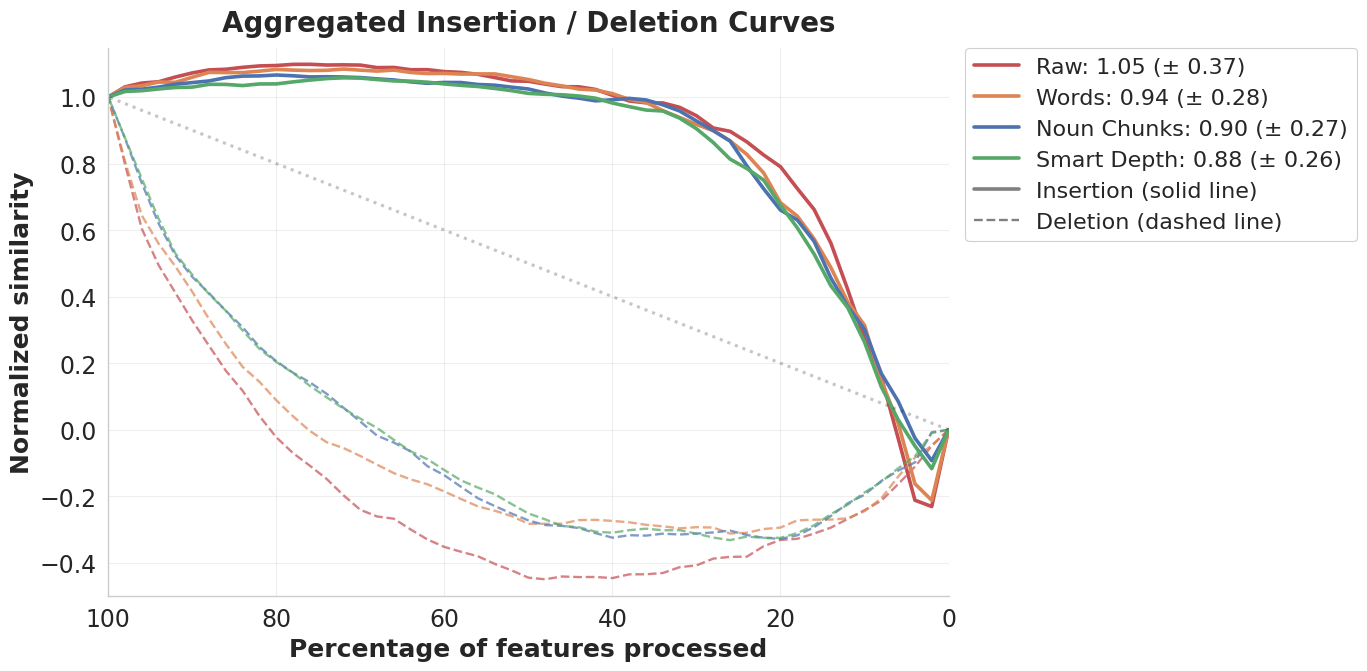}
    \end{minipage}
    \caption{Insertion and Deletion curves. \textbf{Left:} Short Captions ($<10$ words). Semantic grouping outperforms raw tokens on short texts. \textbf{Right:} Long Captions ($>30$ words). On long texts, the raw baseline tracks scattered attention to background noise, slightly inflating its AID score compared to semantic grouping.}
    \label{fig:ins_del_curves}
\end{figure}

In the Figure \ref{fig:ins_del_curves} (Left), semantic grouping strategies
achieve consistently higher AID scores compared to token-level baselines on short captions. \texttt{smart\_depth}
yields the highest AID ($0.79 \pm 0.41$), outperforming the \texttt{raw} strategy ($0.73 \pm 0.34$). Interestingly, on long captions (Figure \ref{fig:ins_del_curves}, Right), the \texttt{raw} strategy achieves slightly higher AID scores, as it may track the model's scattered attention to background noise and adverbs, which semantic grouping intentionally abstracts away.

\section{Qualitative Experiments}

\subsection{Parsing Enables the Understanding of Complex Medical Concepts}

\begin{figure}[htbp]
    \centering
    \begin{minipage}{0.32\linewidth}
        \centering
        \includegraphics[width=\linewidth]{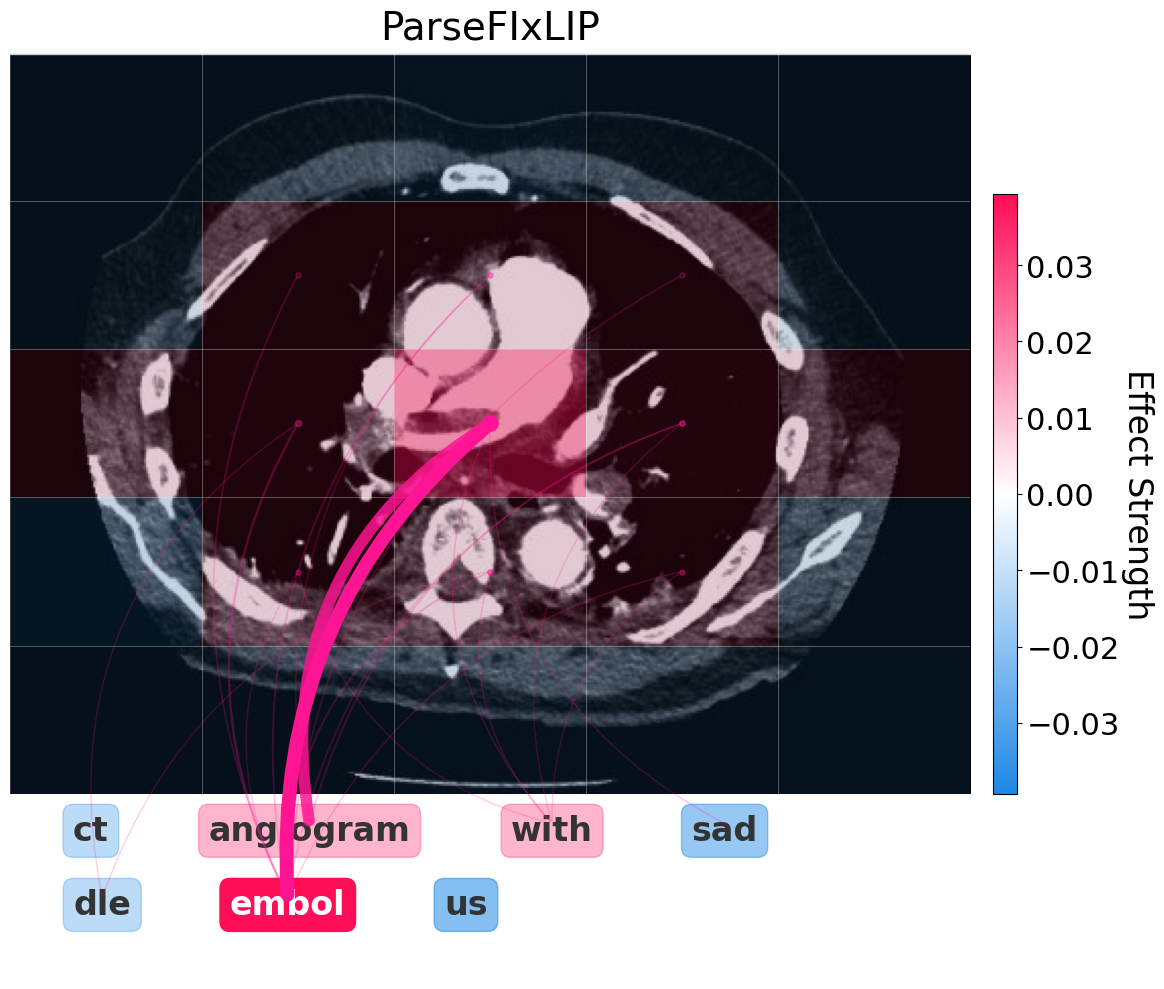}
    \end{minipage}
    \hfill
    \begin{minipage}{0.32\linewidth}
        \centering
        \includegraphics[width=\linewidth]{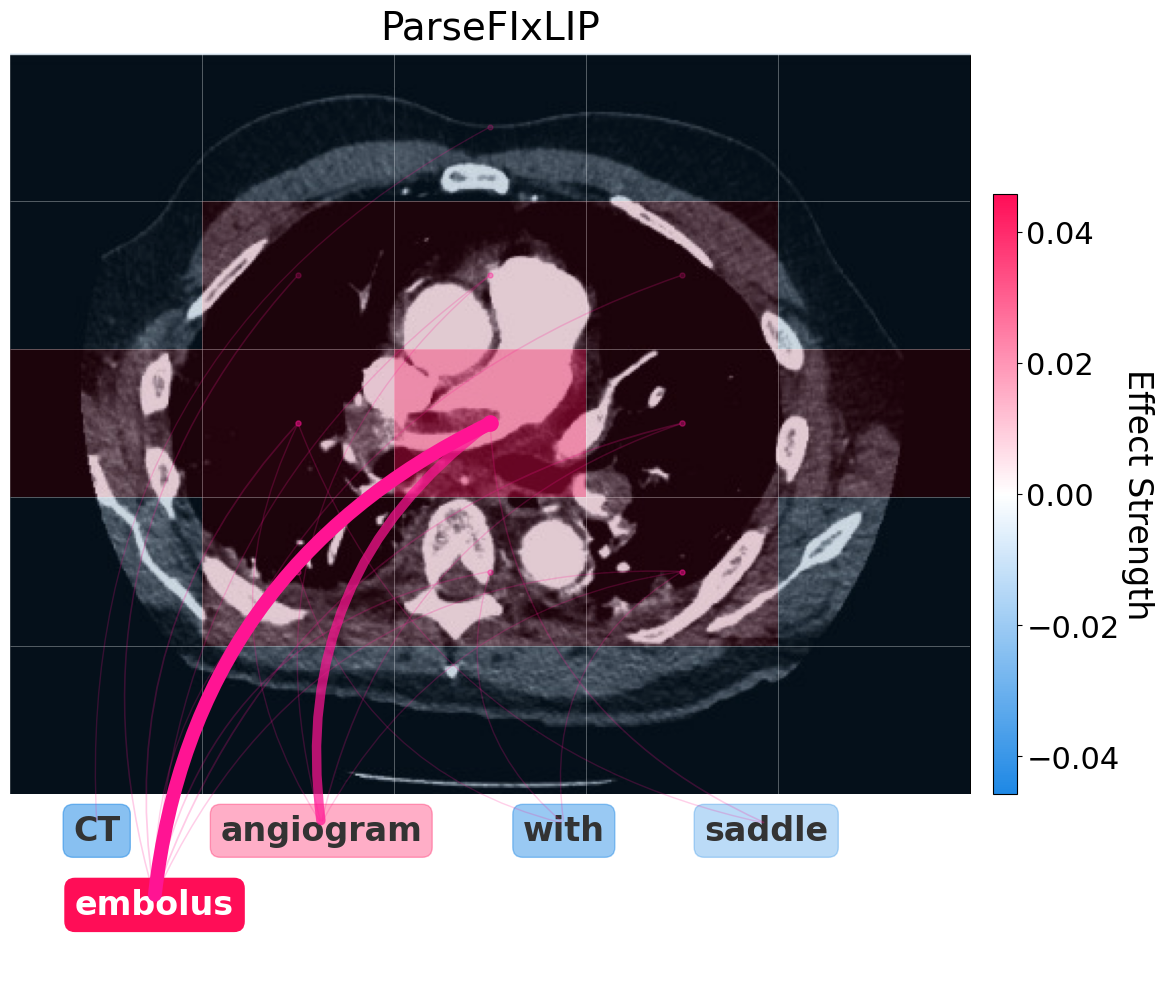}
    \end{minipage}
    \hfill
    \begin{minipage}{0.32\linewidth}
        \centering
        \includegraphics[width=\linewidth]{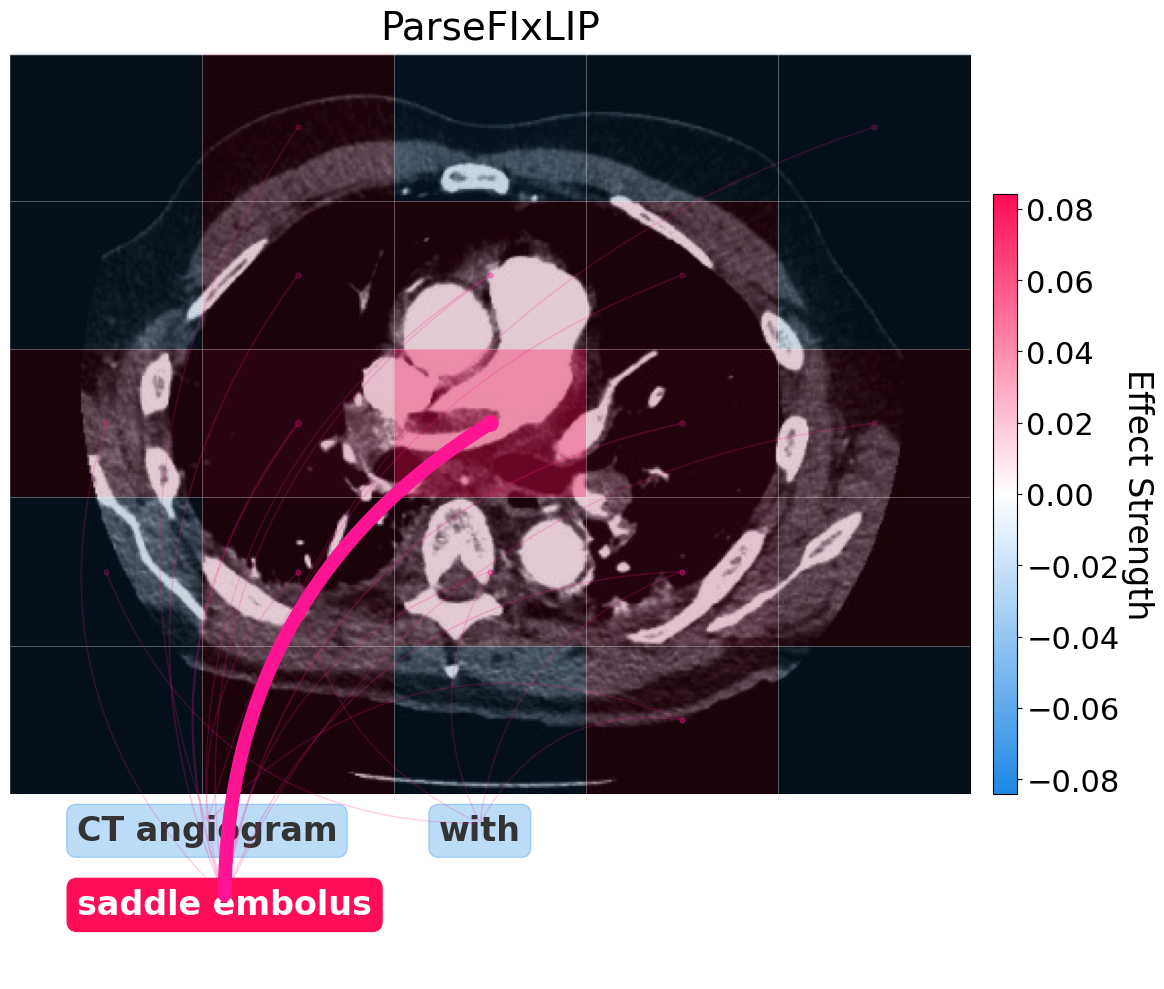}
    \end{minipage}
    \caption{Qualitative comparison of explanations. \textbf{Left:} \texttt{raw} – Concept fragmented, noisy explanation. \textbf{Middle:} \texttt{words} – Subwords merged into words. \textbf{Right:} \texttt{smart\_depth} – ``saddle embolus'' as one concept, a clean explanation.}
    \label{fig:qualitative_saddle}
\end{figure}

In this section, we investigate whether Tree-Gram Parsing mitigates the key limitation of standard VLM explanations, namely the fragmentation of medical concepts caused by tokenization. Figure \ref{fig:qualitative_saddle} illustrates the progression of interpretability across different parsing strategies for the caption ``CT angiogram with saddle embolus''. While our primary qualitative analysis focuses on cross-modal interactions, ParseFIxLIP naturally yields clearer visualizations of main effects (standalone feature attribution). The visualizations show both cross-modal interaction strengths and the main effects of single-modality players.

In the \texttt{raw} strategy, the medical term ``saddle embolus'' is split into arbitrary tokens (e.g., ``sad'', ``dle'', ``embol'', ``us''). The explanation is noisy, with importance distributed across meaningless fragments and significant weight given to the general modality ``angiogram'' rather than the specific pathology ``saddle embolus''. The \texttt{words} strategy merges subwords into recognizable terms. However, it splits similarity score treating ``saddle'' and ``embolus'' as separate players, still highlighting general term ``angiogram''. In contrast, our proposed semantic parsing strategy, \texttt{smart\_depth} successfully groups these tokens into a coherent clinical concept. The parser identifies ``saddle embolus'' as a single entity. Consequently, the explanation concentrates mostly on this unified concept and results in the best quantitative metrics. 

\begin{figure}[htbp]
    \centering
    \begin{minipage}{0.42\linewidth}
        \centering
        \includegraphics[width=\linewidth]{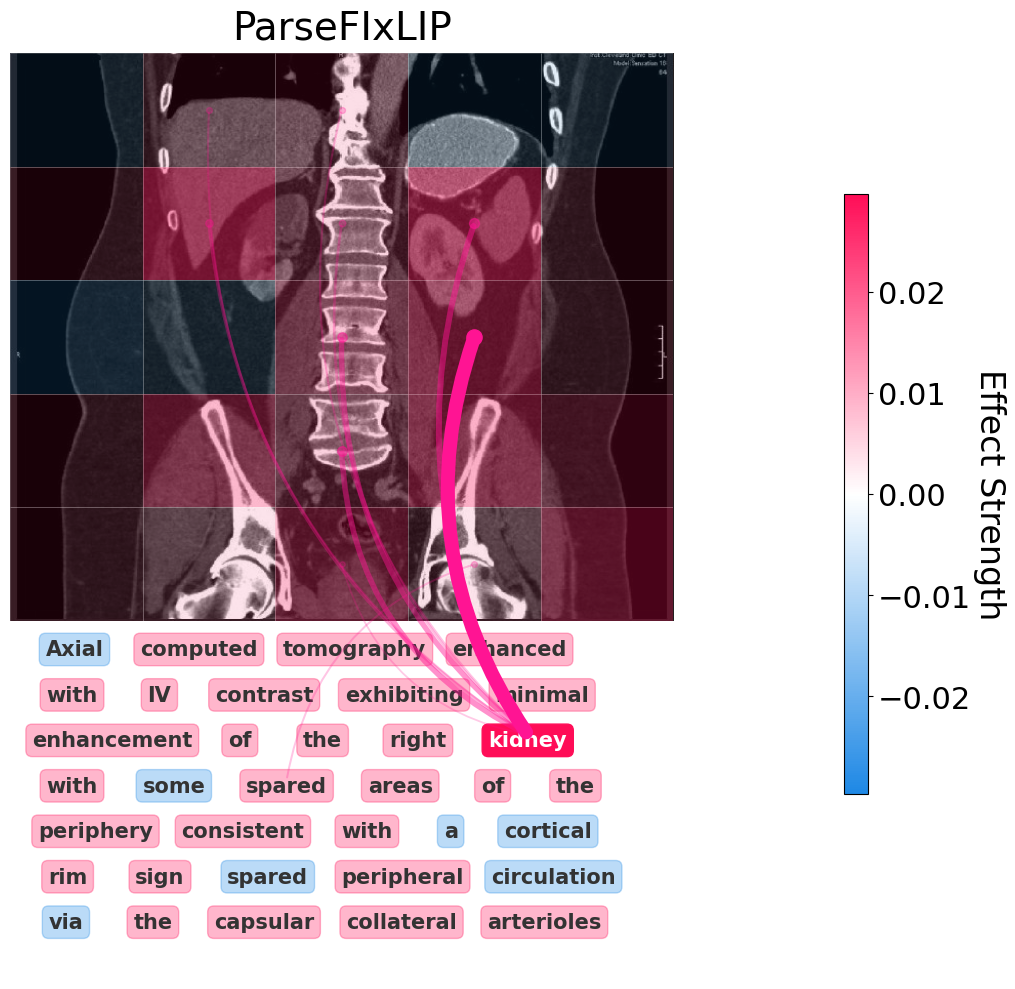}
    \end{minipage}
    \hfill
    \begin{minipage}{0.42\linewidth}
        \centering
        \includegraphics[width=\linewidth]{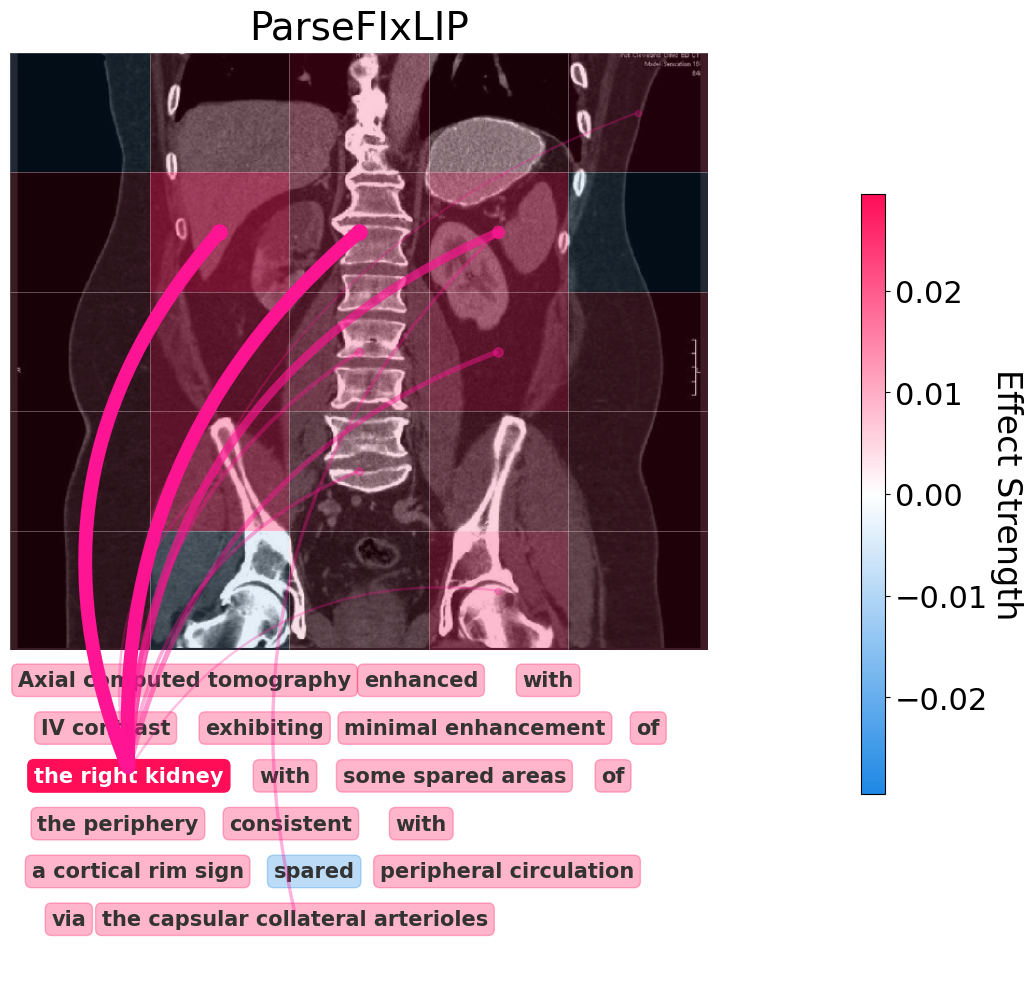}
    \end{minipage}
    \caption{Visual explanation comparison for the long caption containing ``the right kidney''. \textbf{Left:} \texttt{words} strategy -- Fragmentation leads to the inability to distinguish between the kidneys. \textbf{Right:} \texttt{noun\_chunks} -- Semantic grouping unifies these tokens, assigning importance directly to the right kidney.}
    \label{fig:kidney}
\end{figure}

This capability extends effectively to longer and more complex phrases. Figure \ref{fig:kidney} showcases the interaction analysis for the long caption containing the phrase ``the right kidney''. The \texttt{words} strategy fragments the concept into single words, leading to the inability to distinguish between the left and right kidney. Conversely, \texttt{noun\_chunks} processes the entire phrase as a single unit, assigning interaction importance directly to the observed kidney and its anatomical view.

\begin{figure}[htbp]
    \centering
    \begin{minipage}{0.4\linewidth}
        \centering
        \includegraphics[width=\linewidth]{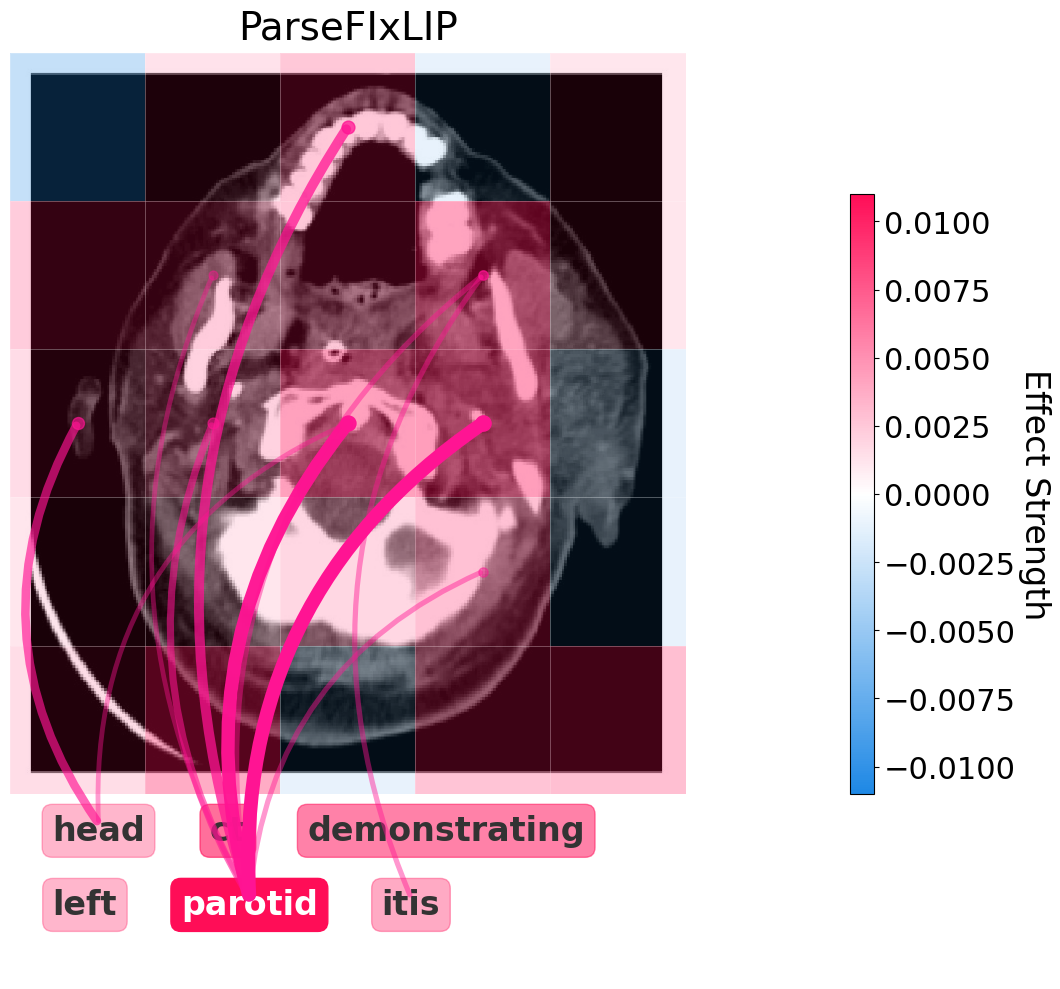}
    \end{minipage}
    \hfill
    \begin{minipage}{0.4\linewidth}
        \centering
        \includegraphics[width=\linewidth]{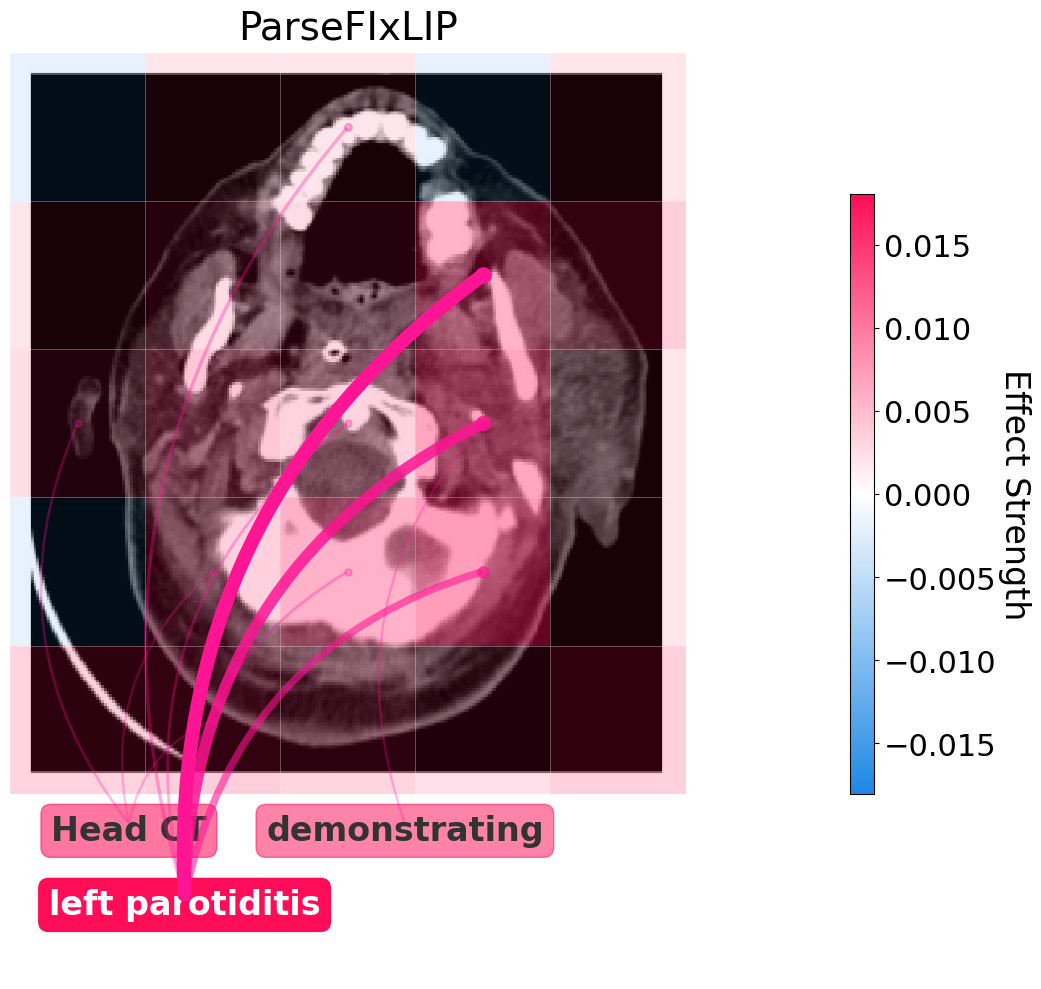}
    \end{minipage}
    \caption{Visual explanation comparison for the caption ``Head CT demonstrating left parotiditis''. 
    \textbf{Left:} \texttt{raw} -- Fragmenting the concept into ``left'', ``parotid'', and ``itis'' results in scattered attention. 
    \textbf{Right:} \texttt{smart\_depth} -- Processing ``left parotiditis'' as a single semantic unit yields a precise explanation. Semantic grouping eliminates the noise seen with fragmented tokens.}
    \label{fig:parotiditis_qualitative}
\end{figure}

\begin{figure}[htbp]
    \centering
    \begin{minipage}{0.4\linewidth}
        \centering
        \includegraphics[width=\linewidth]{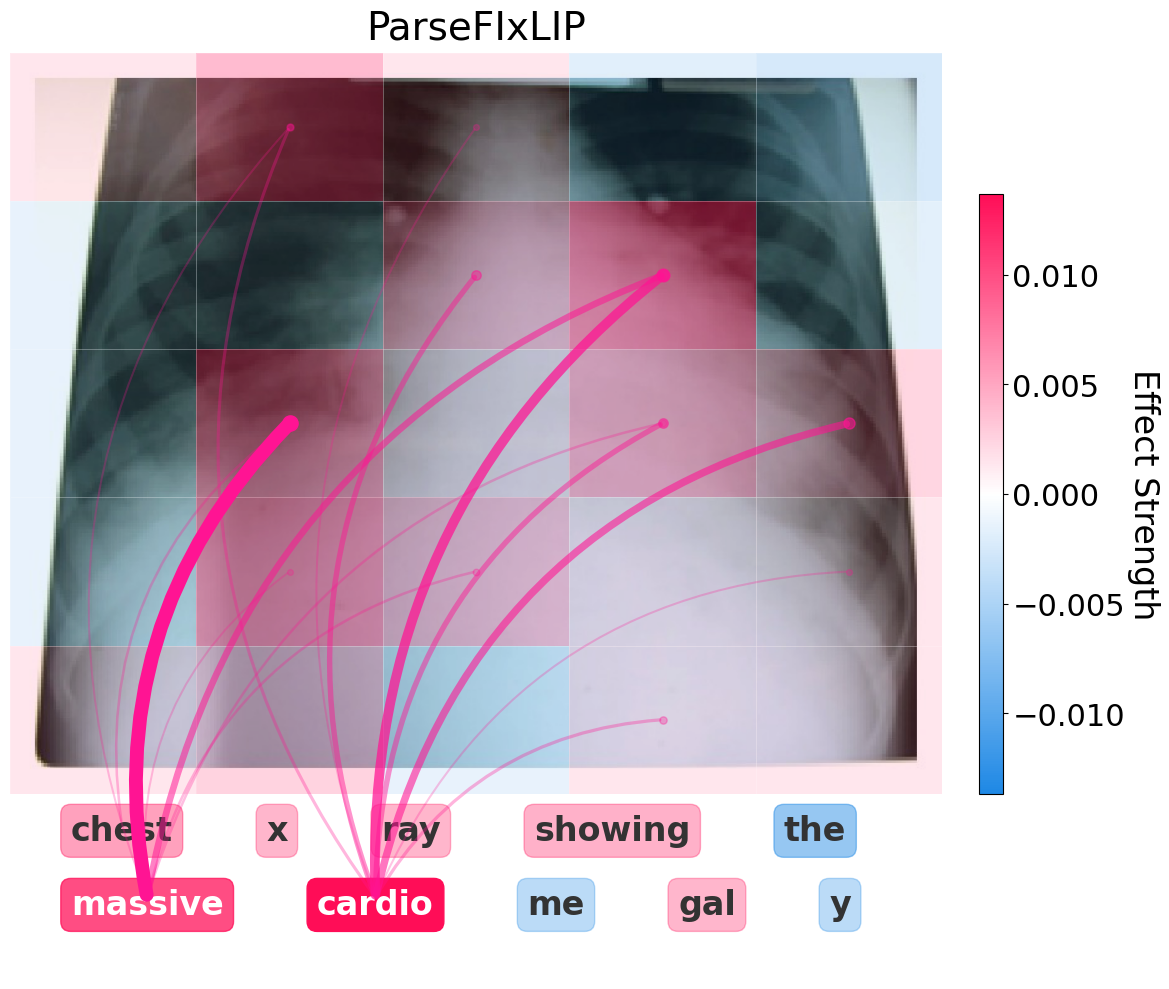}
    \end{minipage}
    \hfill
    \begin{minipage}{0.4\linewidth}
        \centering
        \includegraphics[width=\linewidth]{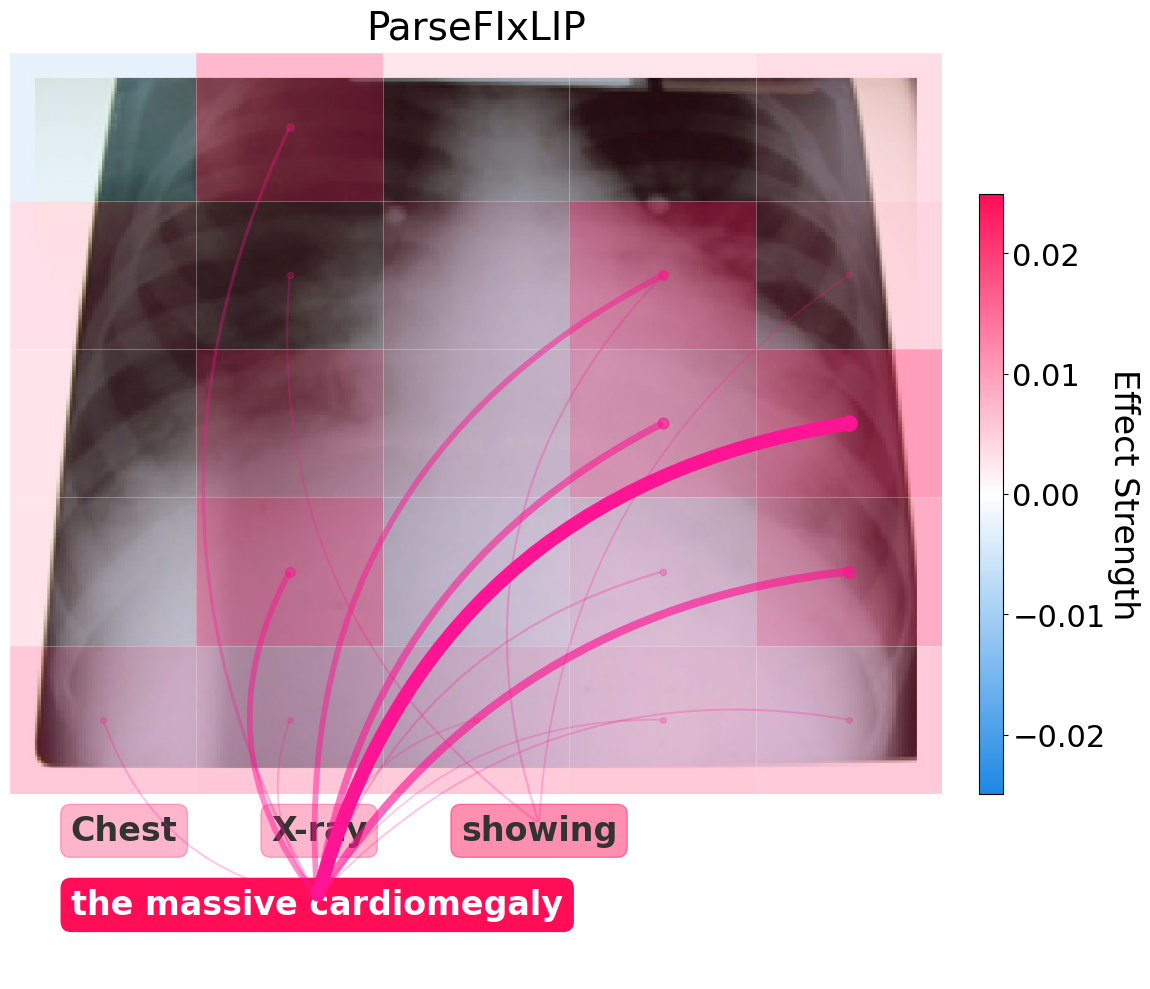}
    \end{minipage}
    \caption{Visual explanation comparison for ``massive cardiomegaly''. \textbf{Left:} \texttt{raw} -- Fragmentation of the medical term generates significant visual noise outside the heart region. \textbf{Right:} \texttt{smart\_depth} –- Processing the entire phrase as a single unit eliminates noise.}
    \label{fig:cardiomegaly_comparison_appendix}
\end{figure}

\begin{figure}[htbp]
    \centering
    \begin{minipage}{0.49\linewidth}
        \centering
        \includegraphics[width=\linewidth]{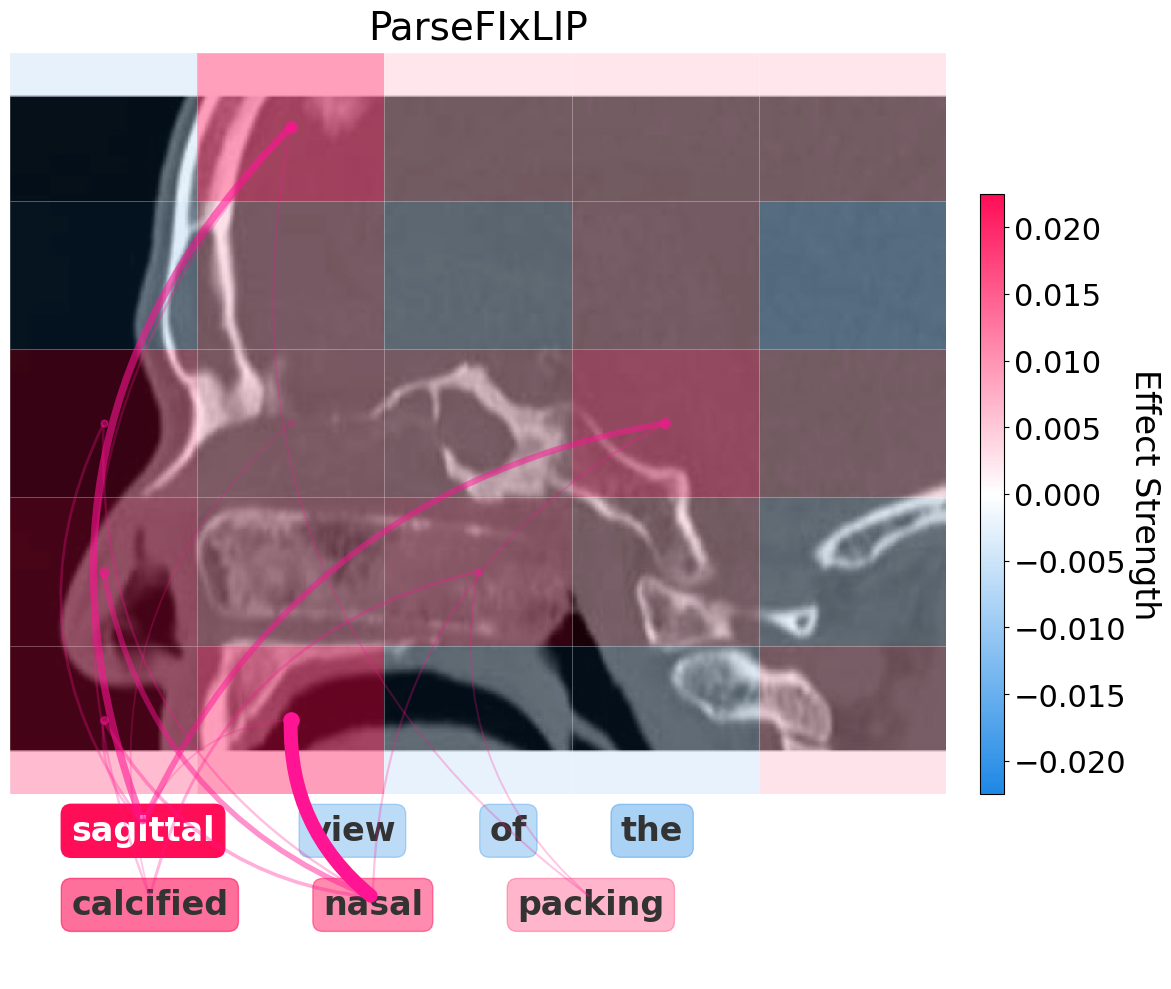} 
    \end{minipage}
    \hfill
    \begin{minipage}{0.49\linewidth}
        \centering
        \includegraphics[width=\linewidth]{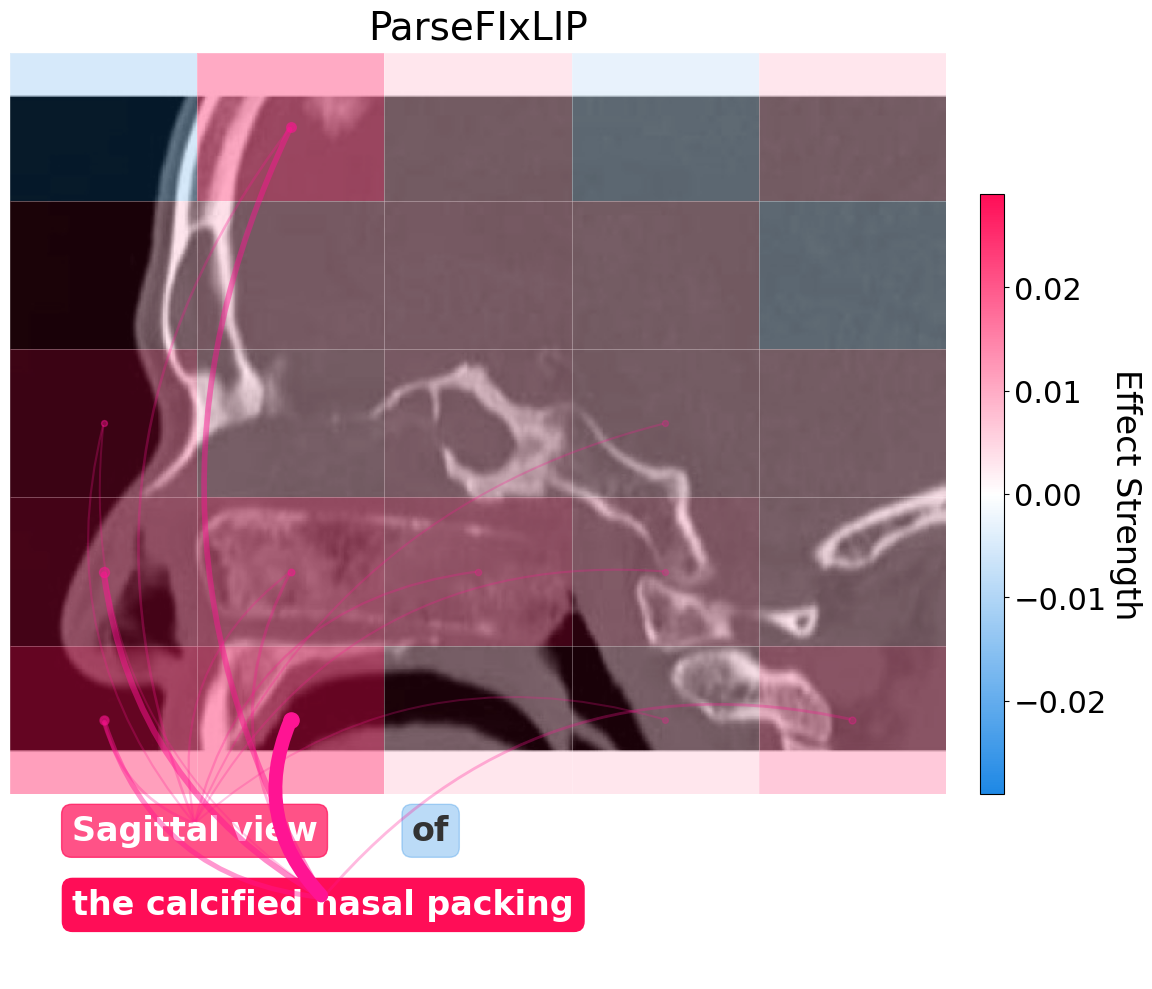} 
    \end{minipage}
    \caption{Visual explanation comparison for ``Sagittal view of the calcified nasal packing''. \textbf{Left:} \texttt{raw} -- Noisy distribution of importance. \textbf{Right:} \texttt{smart\_depth} -- Clear semantic grounding where relevant image parts receive positive attribution.}
    \label{fig:main_effects_sagittal}
\end{figure}

We further illustrate the capability of our \texttt{smart\_depth} strategy to capture complex, multi-word medical concepts compared to the fragmented baseline (Figures \ref{fig:parotiditis_qualitative}, \ref{fig:cardiomegaly_comparison_appendix} and \ref{fig:main_effects_sagittal}).

\begin{figure}[htbp]
    \centering
    \includegraphics[width=0.65\linewidth]{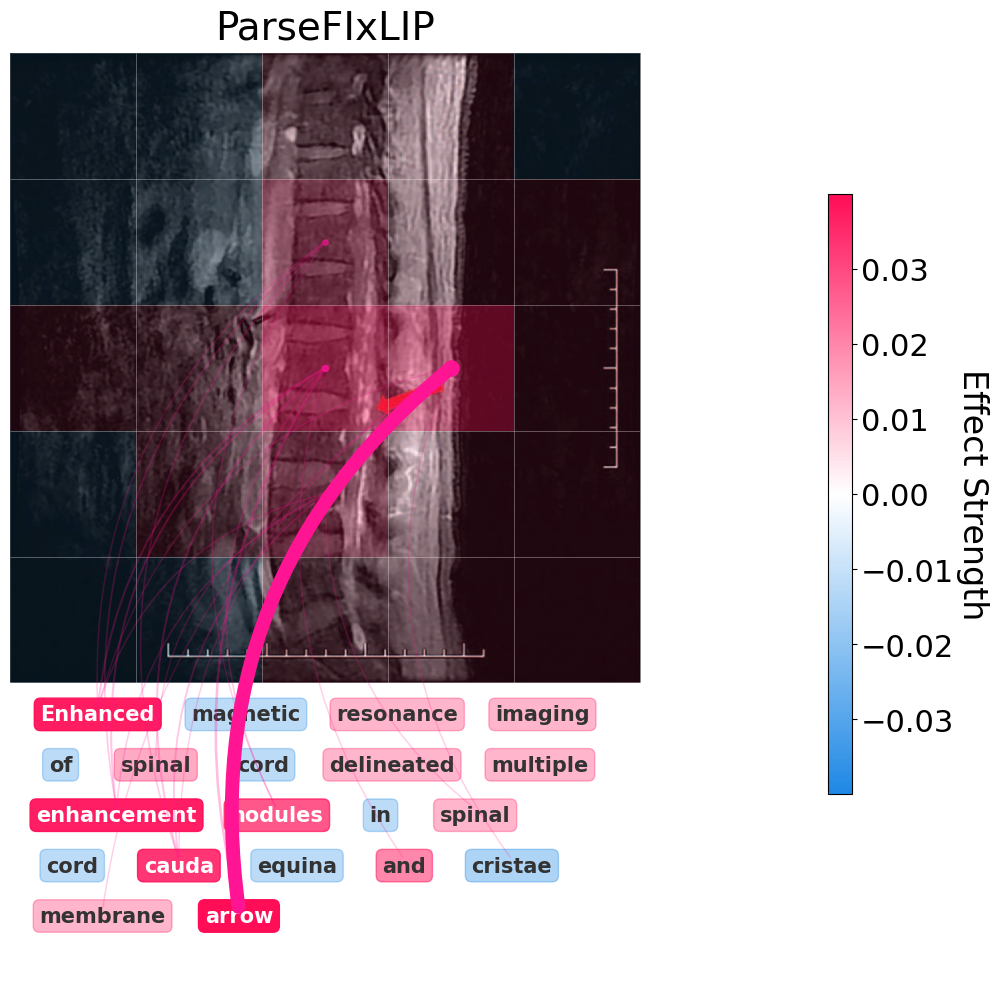} 
    \caption{Visual grounding of explicit annotations. The word ``arrow'' is connected to the red graphical marker pointing towards the spinal pathology.}
    \label{fig:main_effects_arrow}
\end{figure}

Radiological images frequently contain explicit annotations (arrows, rulers, text, like in the Figure \ref{fig:main_effects_arrow}) which pose a unique challenge. Our framework connects these straightforward, not interesting from a medical perspective, interactions. For instance, the word ``arrow'' is heavily connected to the red graphical marker pointing towards the spinal pathology. Some other seemingly important relations can be omitted.

\newpage

\subsection{Image transformation test}

We investigate the robustness of BiomedCLIP and our parsing strategy to spatial transformations such as upside down rotation. The baseline normal image yields a model output similarity of 0.4587, whereas the transformed version drops to 0.3061. This performance degradation is likely because the model was trained on medical images always taken from a standardized position, thus the model does not recognize the concept of the right lobe in the rotated image and the main contributor are the numbers. Grouping tokens strengthens ``the right lobe'' score for normal view. The model demonstrates robustness to transformations regarding printed numbers and symbols, but fails to maintain orientation-awareness for anatomical structures.

\begin{figure}[htbp]
    \centering
    \begin{minipage}{0.49\linewidth}
        \centering
        \includegraphics[width=\linewidth]{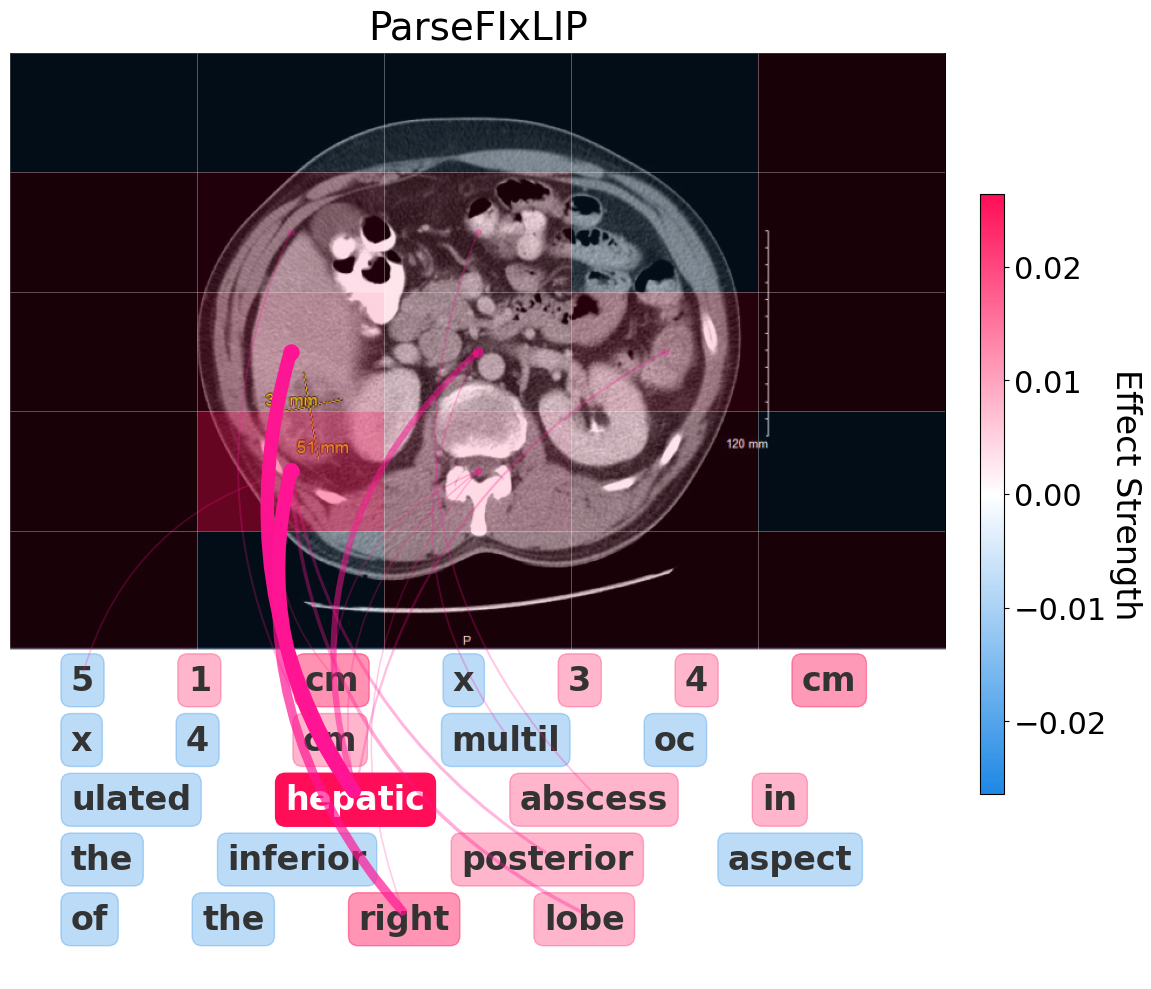}
    \end{minipage}
    \hfill
    \begin{minipage}{0.49\linewidth}
        \centering
        \includegraphics[width=\linewidth]{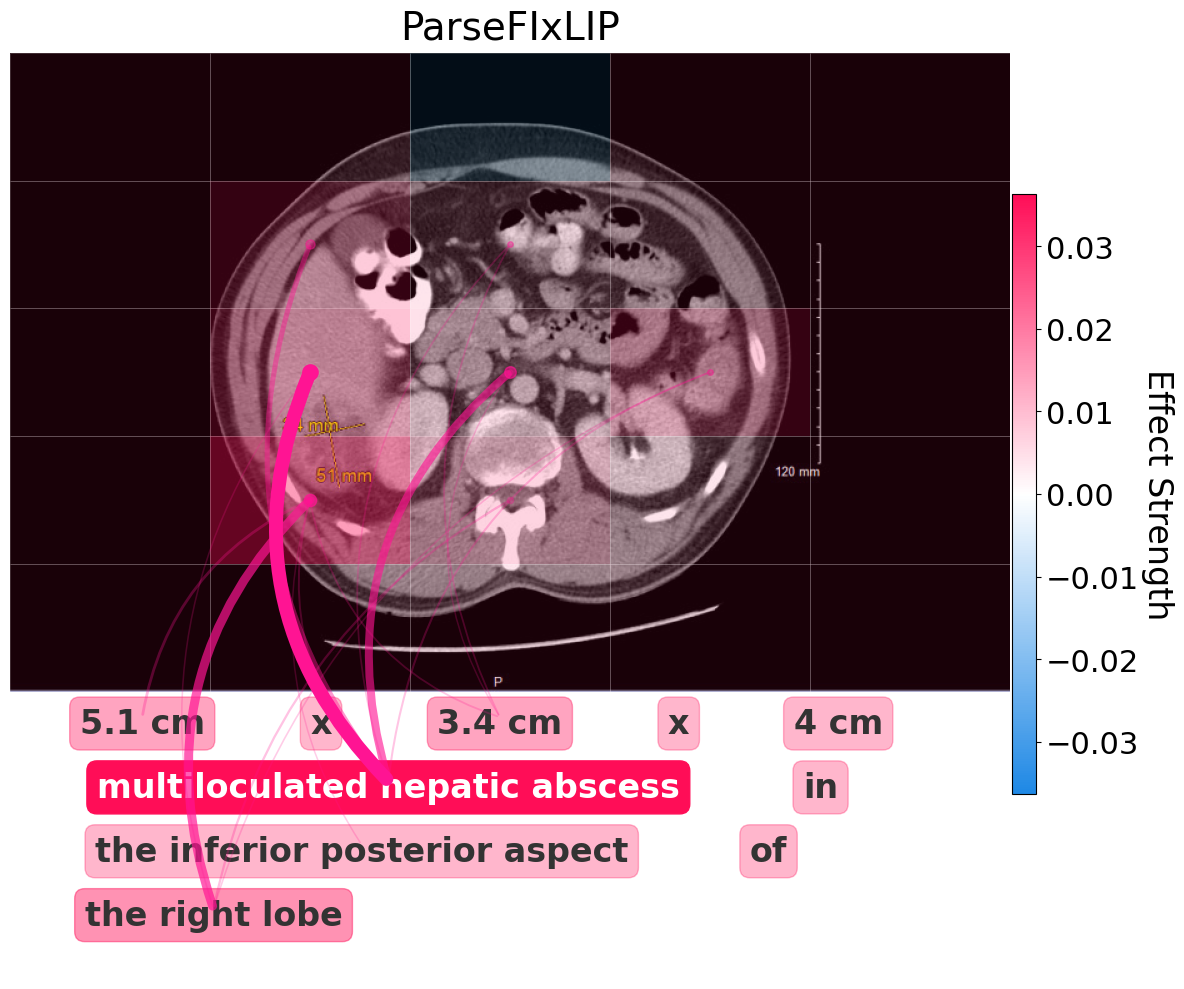}
    \end{minipage}
    \caption{Image transformation comparison. \textbf{Left:} Normal \texttt{raw}. \textbf{Right:} Normal \texttt{smart\_depth}.}
    \label{fig:normal_liver}
\end{figure}

\begin{figure}[htbp]
    \centering
    \begin{minipage}{0.49\linewidth}
        \centering
        \includegraphics[width=\linewidth]{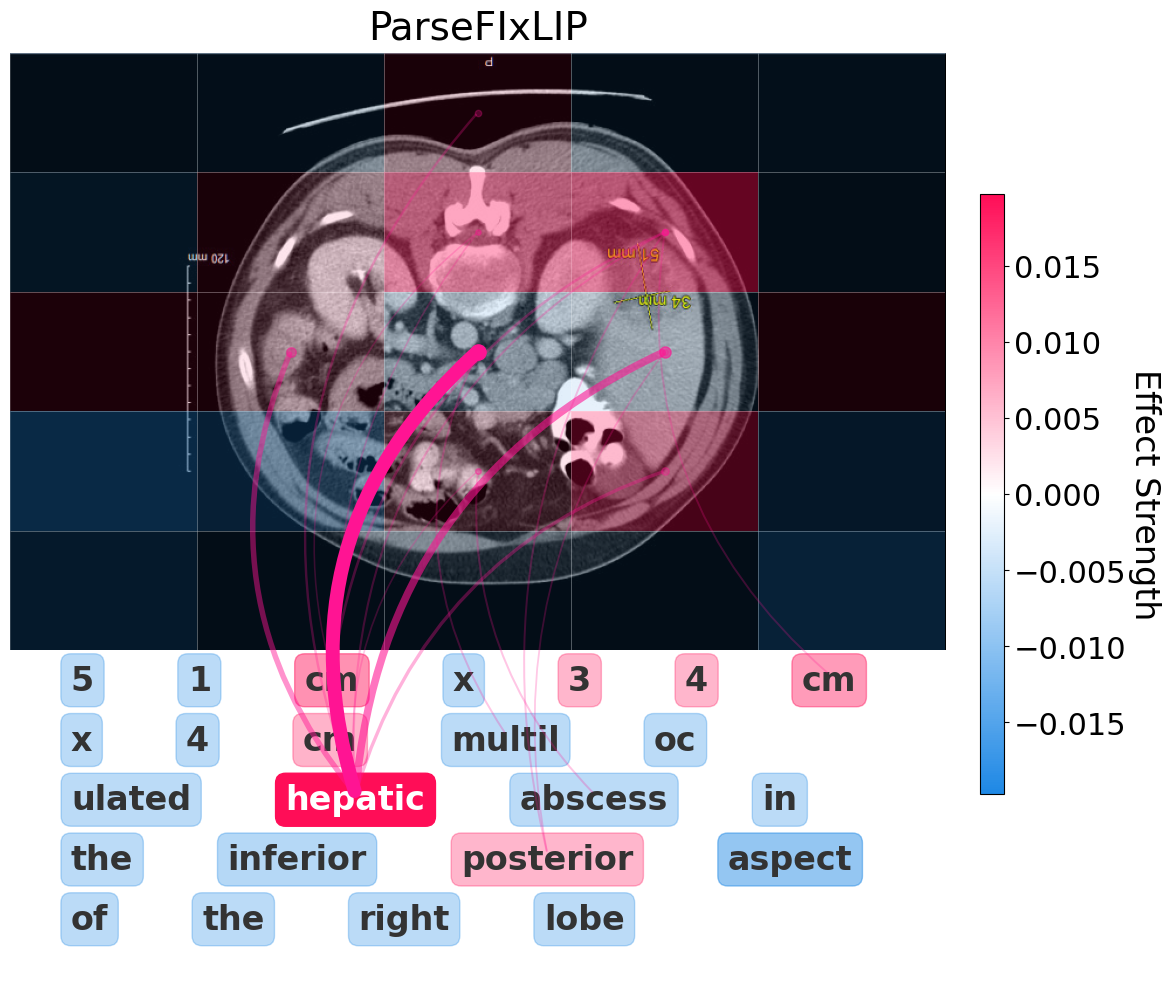}
    \end{minipage}
    \hfill
    \begin{minipage}{0.49\linewidth}
        \centering
        \includegraphics[width=\linewidth]{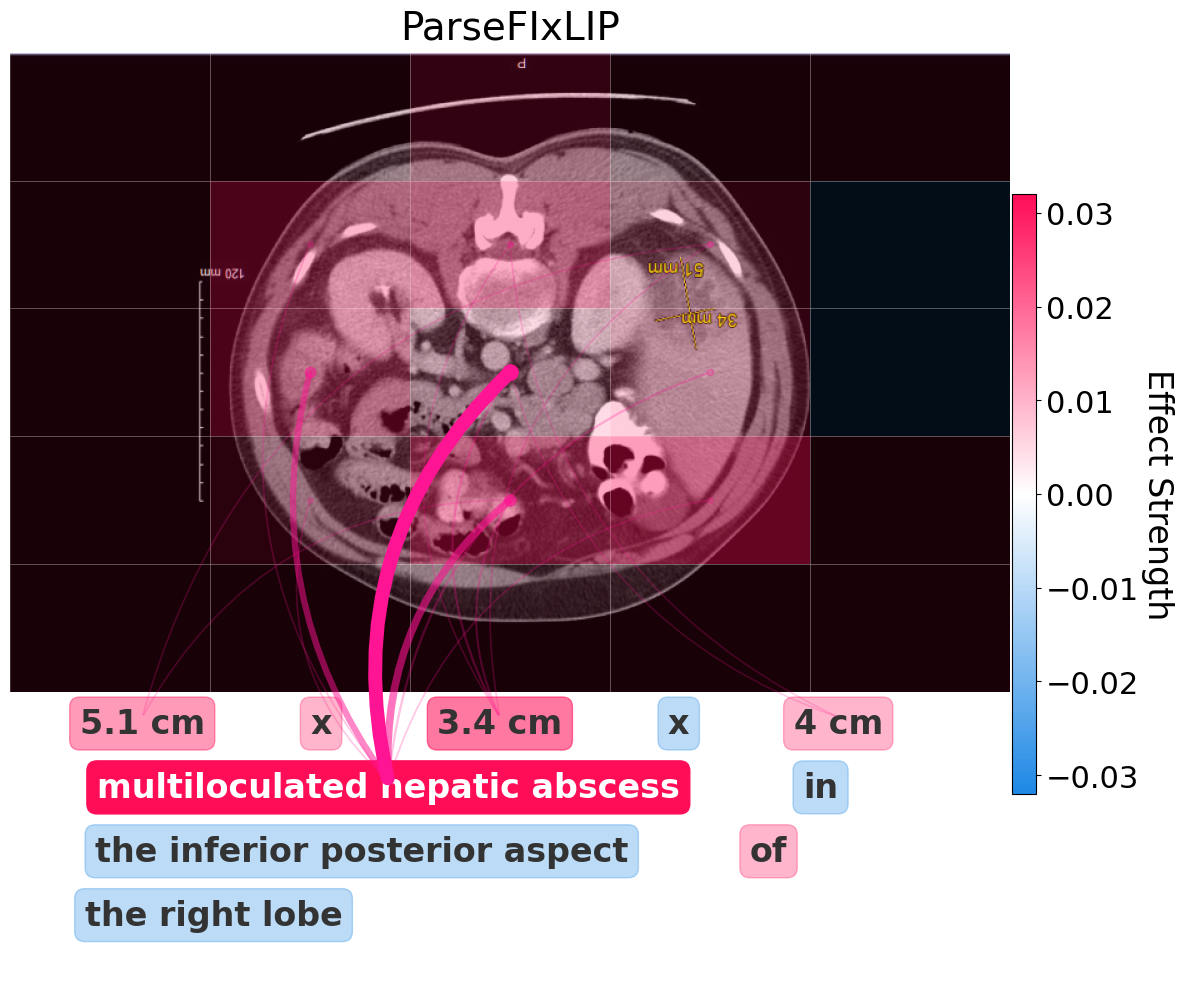}
    \end{minipage}
    \caption{Image transformation comparison. \textbf{Left:} Upside down \texttt{raw}. \textbf{Right:} Upside down \texttt{smart\_depth}.}
    \label{fig:upside_down_liver}
\end{figure}

The model exhibits unexpected behavior when processing inverted images. Deducing from the Figures \ref{fig:normal_liver} and \ref{fig:upside_down_liver}, it seems that the model is learning ``multiloculated hepatic abscess'' from areas other than the liver. This may be related to ``hepatic'' domination seen in the \texttt{raw}, which word regularly appears in the vision context (and can be linked with liver-neighboring organs).

\subsection{False label test}
\label{sec:app_falselabel}

\begin{figure}[htbp]
    \centering
    \begin{minipage}{0.49\linewidth}
        \centering
        \includegraphics[width=\linewidth]{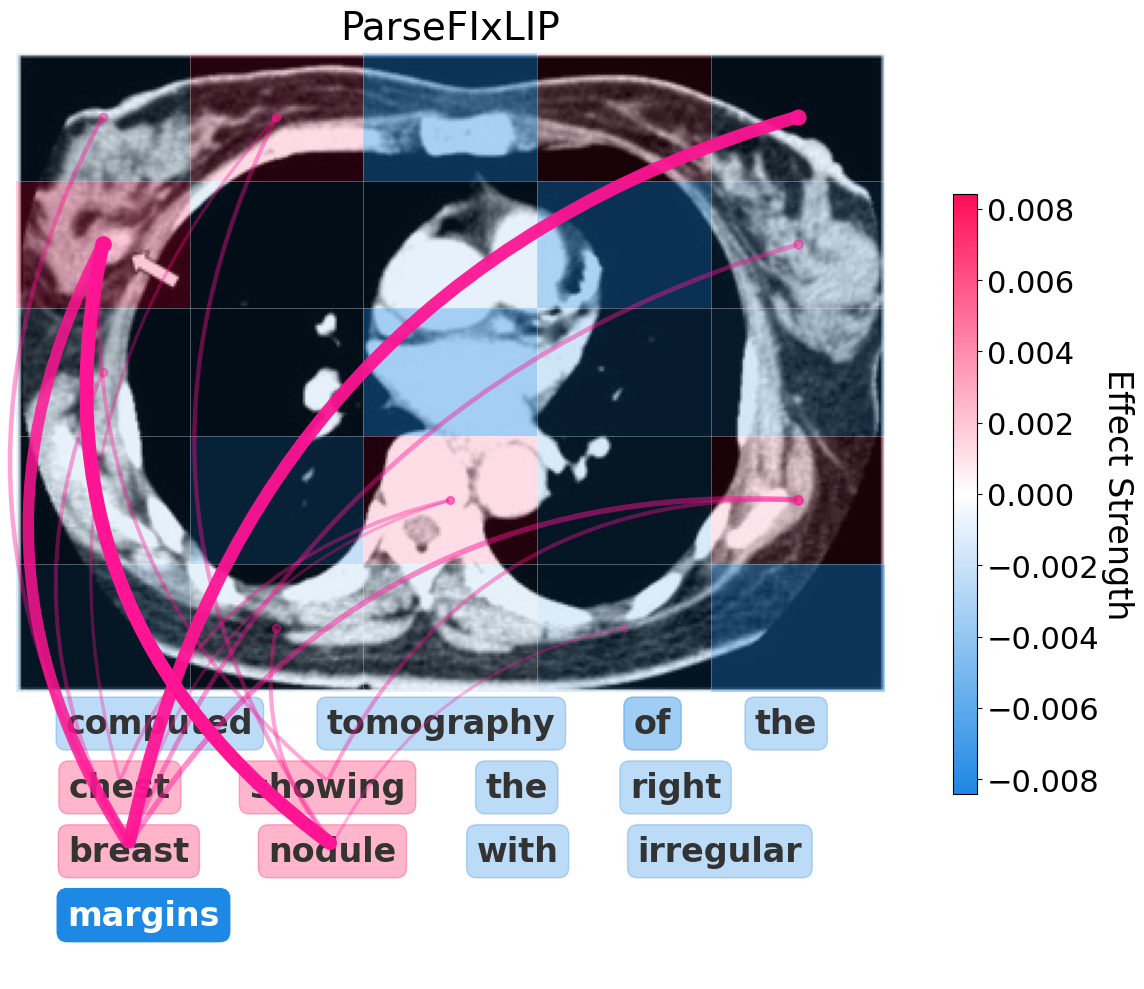}
    \end{minipage}
    \hfill
    \begin{minipage}{0.49\linewidth}
        \centering
        \includegraphics[width=\linewidth]{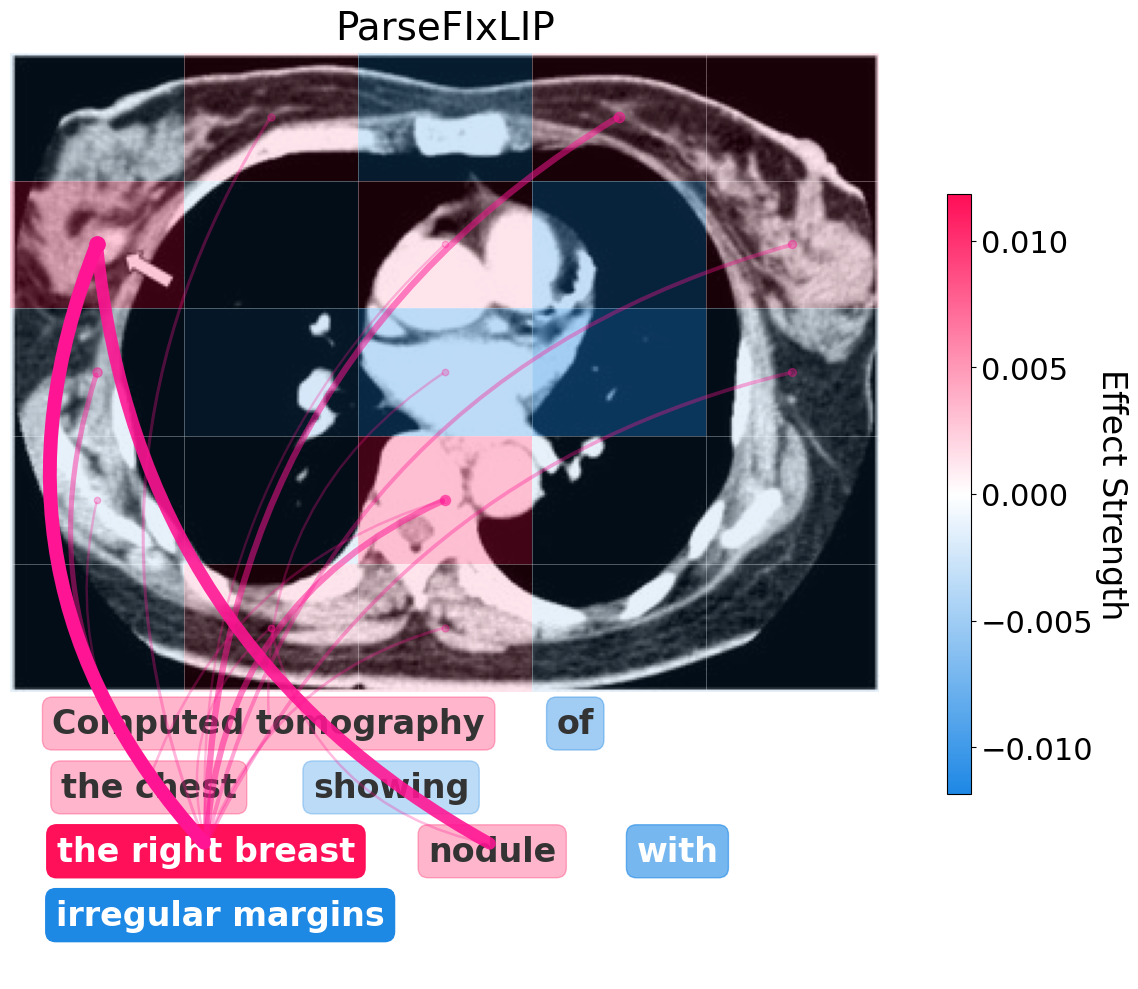}
    \end{minipage}
    \caption{True label. \textbf{Left:} \texttt{raw}. \textbf{Right:} \texttt{smart\_depth}.}
    \label{fig:true_label_breast}
\end{figure}

\begin{figure}[htbp]
    \centering
    \begin{minipage}{0.49\linewidth}
        \centering
        \includegraphics[width=\linewidth]{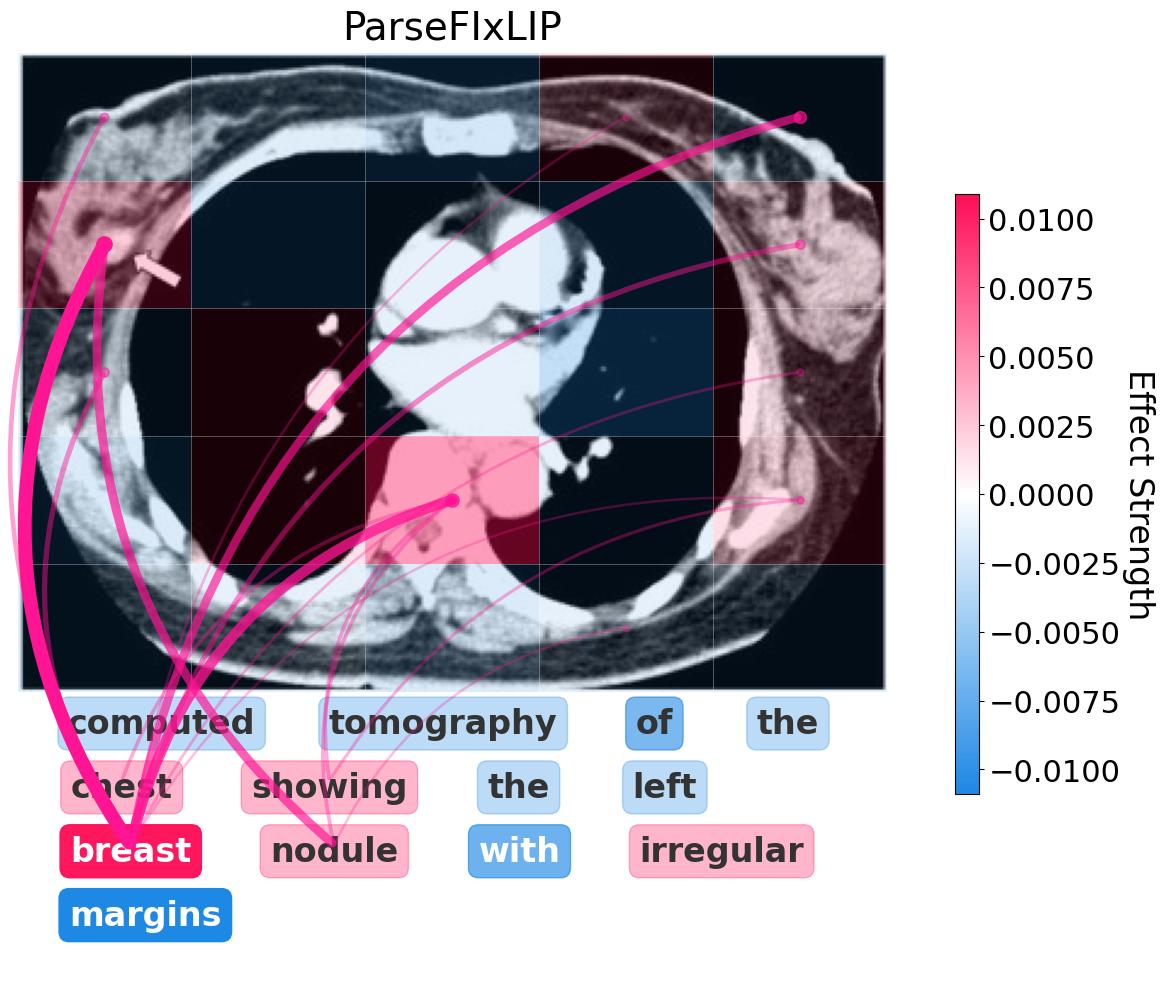}
    \end{minipage}
    \hfill
    \begin{minipage}{0.49\linewidth}
        \centering
        \includegraphics[width=\linewidth]{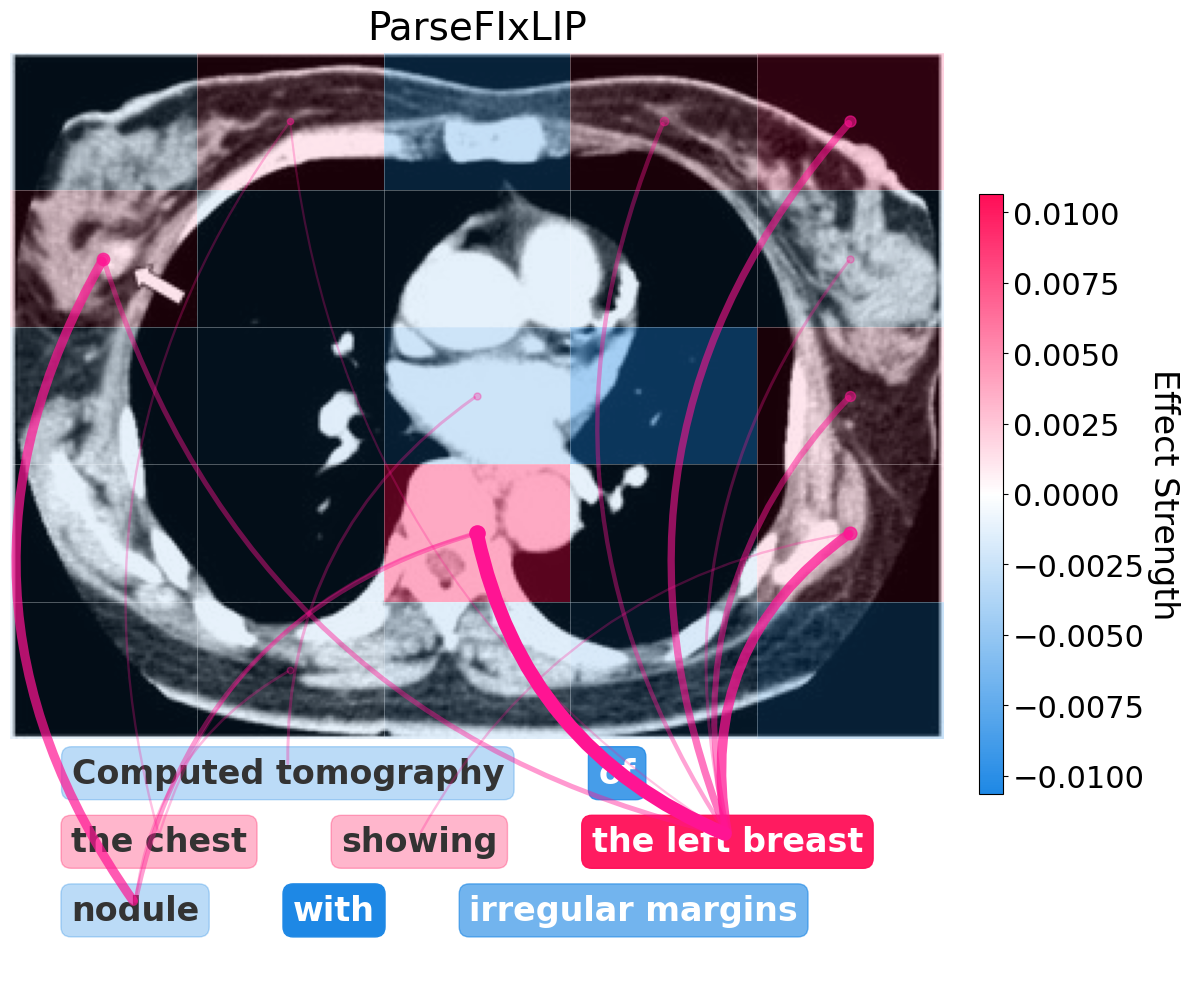}
    \end{minipage}
    \caption{False label. \textbf{Left:} \texttt{raw}. \textbf{Right:} \texttt{smart\_depth}.}
    \label{fig:false_label_breast}
\end{figure}

\textbf{Can the model detect left/right change in the text?} The model can distinguish between the left and the right breast (Figures \ref{fig:true_label_breast}, \ref{fig:false_label_breast}), probably considering its placement in the image. The breast position on the image is meaningful for selecting the correct breast, not the arrow. Connecting the tokens together changes the part of the image indicated by breast — breast appearing always with the correct left/right indicator is linked with the image part correctly, not as the right. 

The model fails in recognizing the correct breast in the transformed image (not surprising, as breasts are quite similar and model is not transformation-robust). A notable limitation observed in this experiment is that the Tree-Gram Parsing algorithm separated the ``breast'' token from the ``nodule'' token (probably due to being separate nouns). It would be interesting to see if the model is more likely to classify such grouped chunk to the left breast (false label) or the nodule (detected correctly on the right breast). Interestingly, both the nodule and the arrow pointing to it are visible in the image. They are on separate grids, but all the experiments show that the arrow is more meaningful for nodule detection in the image than the nodule itself.

\section{Conclusions}
One of the goals of our work is to faithfully explain the inner workings of BiomedCLIP. To demonstrate this capability and evaluate our method, we conducted several qualitative stress tests. Through these tests, we observe that BiomedCLIP itself is not robust to certain conditions, which our explanations successfully capture. \textbf{Transformation Fragility.} Because medical datasets contain standardized images, BiomedCLIP severely lacks robustness to spatial transformations. When images are rotated (Figure \ref{fig:upside_down_liver}), the model's similarity score drops drastically. ParseFIxLIP faithfully reflects this confusion: the interaction weight shifts away from the misplaced anatomy and heavily focuses on transformation-robust artifacts, such as printed numbers or text symbols. \textbf{Shortcut Learning via Annotations.} Radiological images frequently contain explicit markers like arrows or text. Our explanations reveal that BiomedCLIP frequently assigns disproportionately high interaction weights to these explicit pointers rather than the actual pathology. This highlights a critical warning: the model often learns to ``cheat'' by looking at human annotations, meaning interpretations of BiomedCLIP outputs require extreme caution. \textbf{Parser Limitations.} While Tree-Gram Parsing effectively unifies clinical concepts, it is fundamentally constrained by the underlying NLP parser (spaCy). As observed in our false label tests (Section \ref{sec:app_falselabel}), the parser occasionally separates related concepts (e.g., decoupling ``breast'' from ``nodule'').

To conclude, adding Tree-Gram Parsing to the FIxLIP algorithm allows us to understand more about how tokens influence the model results and find out the synergy of the grouped tokens. Quantitative evaluation shows that our parsing-based approach achieves performance higher and more stable than the raw token-based FIxLIP, efficiently neutralizing the curse of dimensionality on complex reports. Qualitative analysis demonstrates that our dependency-based grouping significantly mitigates the fragmentation issues inherent in the baseline method, offering more intuitive and clinically relevant insights into the model’s decision-making process. Crucially, ParseFIxLIP exposes BiomedCLIP's severe limitations, such as spatial transformation fragility and shortcut learning via visual annotations (arrows, numbers).

\section*{Declaration on Generative AI}
During the preparation of this work, the authors used \textit{Gemini} in order to: Grammar and spelling check. After using this tool, the authors reviewed and edited the content as needed and take full responsibility for the publication's content.

\bibliography{references}

@inproceedings{baniecki2025fixlip,
    title     = {Explaining Similarity in Vision-Language Encoders with Weighted Banzhaf Interactions},
    author={Baniecki, Hubert and Muschalik, Maximilian and Fumagalli, Fabian and Hammer, Barbara and H{\"u}llermeier, Eyke and Biecek, Przemyslaw},
    booktitle = {Advances in Neural Information Processing Systems},
    year      = {2025}
}

@inproceedings{simaan2000tree,
  title={Tree-gram parsing: Lexical dependencies and structural relations},
  author={Sima’an, Khalil},
  booktitle={Proceedings of the 38th annual meeting of the association for computational linguistics},
  pages={37--44},
  year={2000}
}

@software{spacy,
  author       = {Honnibal, Matthew and Montani, Ines and Van Landeghem, Sofie and Boyd, Adriane},
  title        = {spaCy: Industrial-strength Natural Language Processing in Python},
  year         = {2020},
  publisher    = {Zenodo},
  doi          = {10.5281/zenodo.1212303},
  url          = {https://spacy.io}
}

@article{zhang2023biomedclip,
  title={A Multimodal Biomedical Foundation Model Trained from Fifteen Million Image--Text Pairs},
  author={Zhang, Sheng and Xu, Yanbo and Usuyama, Naoto and Xu, Hanwen and Bagga, Jaspreet and Tinn, Robert and Preston, Sam and Rao, Rajesh and Wei, Mu and Valluri, Naveen and others},
  journal={NEJM AI},
  volume={2},
  year={2024},
  doi={10.1056/AIoa2400640}
}

@article{ruckert2024rocov2,
  title={Rocov2: Radiology objects in context version 2, an updated multimodal image dataset},
  author={R{\"u}ckert, Johannes and Bloch, Louise and Br{\"u}ngel, Raphael and Idrissi-Yaghir, Ahmad and Sch{\"a}fer, Henning and Schmidt, Cynthia S and Koitka, Sven and Pelka, Obioma and Abacha, Asma Ben and G. Seco de Herrera, Alba and others},
  journal={Scientific Data},
  volume={11},
  number={1},
  pages={688},
  year={2024},
  publisher={Nature Publishing Group UK London}
}

@inproceedings{radford2021learning,
  title={Learning transferable visual models from natural language supervision},
  author={Radford, Alec and Kim, Jong Wook and Hallacy, Chris and Ramesh, Aditya and Goh, Gabriel and Agarwal, Sandhini and Sastry, Girish and Askell, Amanda and Mishkin, Pamela and Clark, Jack and others},
  booktitle={International Conference on Machine Learning},
  pages={8748--8763},
  year={2021},
  organization={PMLR}
}

@inproceedings{chefer2021generic,
  title={Generic attention-model explainability for interpreting bi-modal and encoder-decoder transformers},
  author={Chefer, Hila and Gur, Shir and Wolf, Lior},
  booktitle={Proceedings of the IEEE/CVF international conference on computer vision},
  pages={397--406},
  year={2021}
}

@inproceedings{zhao2024grad,
  title={Gradient-based Visual Explanation for Transformer-based {CLIP}},
  author={Zhao, Chenyang and Wang, Kun and Zeng, Xingyu and Zhao, Rui and Chan, Antoni B.},
  booktitle={International Conference on Machine Learning},
  pages={61072--61091},
  year={2024},
  organization={PMLR}
}

@inproceedings{jia2021scaling,
  title={Scaling up visual and vision-language representation learning with noisy text supervision},
  author={Jia, Chao and Yang, Yinfei and Xia, Ye and Chen, Yi-Ting and Parekh, Zarana and Pham, Hieu and Le, Quoc and Sung, Yun-Hsuan and Li, Zhen and Duerig, Tom},
  booktitle={International Conference on Machine Learning},
  pages={4904--4916},
  year={2021},
  organization={PMLR}
}

@inproceedings{li2022blip,
  title={Blip: Bootstrapping language-image pre-training for unified vision-language understanding and generation},
  author={Li, Junnan and Li, Dongxu and Xiong, Caiming and Hoi, Steven},
  booktitle={International Conference on Machine Learning},
  pages={12888--12900},
  year={2022},
  organization={PMLR}
}

@article{acosta2022multimodal,
  title={Multimodal biomedical AI},
  author={Acosta, Juan N and Falcone, Guido J and Rajpurkar, Pranav and Topol, Eric J},
  journal={Nature Medicine},
  volume={28},
  number={9},
  pages={1773--1784},
  year={2022},
  publisher={Nature Publishing Group US New York}
}

@article{amann2020explainability,
  title={Explainability for artificial intelligence in healthcare: a multidisciplinary perspective},
  author={Amann, Julia and Blasimme, Alessandro and Vayena, Effy and Frey, Dietmar and Madai, Vince I and {Precise4Q Consortium}},
  journal={BMC medical informatics and decision making},
  volume={20},
  number={1},
  pages={310},
  year={2020},
  publisher={Springer}
}

@inproceedings{bordt2023shapley,
  title={From shapley values to generalized additive models and back},
  author={Bordt, Sebastian and von Luxburg, Ulrike},
  booktitle={International Conference on Artificial Intelligence and Statistics},
  pages={709--745},
  year={2023},
  organization={PMLR}
}

@inproceedings{fumagalli2023shapiq,
  title={{SHAP-IQ}: Unified Approximation of any-order Shapley Interactions},
  author={Fumagalli, Fabian and Muschalik, Maximilian and Kolpaczki, Patrick and H{\"u}llermeier, Eyke and Hammer, Barbara},
  booktitle={Advances in Neural Information Processing Systems},
  year={2023}
}

@article{pelegrina2023additive,
  title={A k-additive Choquet integral-based approach to approximate the SHAP values for local interpretability in machine learning},
  author={Pelegrina, Guilherme Dean and Duarte, Leonardo Tomazeli and Grabisch, Michel},
  journal={Artificial Intelligence},
  volume={325},
  pages={104014},
  year={2023},
  publisher={Elsevier}
}

@article{marichal2011weighted,
  title={Weighted Banzhaf power and interaction indexes through weighted approximations of games},
  author={Marichal, Jean-Luc and Mathonet, Pierre},
  journal={European Journal of Operational Research},
  volume={211},
  number={2},
  pages={352--358},
  year={2011},
  publisher={Elsevier}
}

@inproceedings{covert2021improving,
  title={Improving kernelshap: Practical shapley value estimation using linear regression},
  author={Covert, Ian and Lee, Su-In},
  booktitle={International Conference on Artificial Intelligence and Statistics},
  pages={3457--3465},
  year={2021},
  organization={PMLR}
}

@book{hastie2009elements,
  title={The Elements of Statistical Learning: Data Mining, Inference, and Prediction},
  author={Hastie, Trevor and Tibshirani, Robert and Friedman, Jerome},
  year={2009},
  publisher={Springer Science \& Business Media}
}

@article{moor2023foundation,
  title={Foundation models for generalist medical artificial intelligence},
  author={Moor, Michael and Banerjee, Oishi and Abad, Zahra Shakeri Hossein and Krumholz, Harlan M and Leskovec, Jure and Topol, Eric J and Rajpurkar, Pranav},
  journal={Nature},
  volume={616},
  number={7956},
  pages={259--265},
  year={2023},
  publisher={Nature Publishing Group UK London}
}

@article{covert2021explaining,
  title={Explaining by removing: A unified framework for model explanation},
  author={Covert, Ian and Lundberg, Scott and Lee, Su-In},
  journal={Journal of Machine Learning Research},
  volume={22},
  number={209},
  pages={1--90},
  year={2021}
}

@article{mielke2021between,
  title={Between words and characters: A brief history of open-vocabulary modeling and tokenization in NLP},
  author={Mielke, Sabrina J and Alyafeai, Zaid and Salesky, Elizabeth and Raffel, Colin and Dey, Manan and Gall{\'e}, Matthias and Raja, Arun and Si, Chenglei and Lee, Wilson Y and Sagot, Beno{\^\i}t and others},
  journal={arXiv preprint arXiv:2112.10508},
  year={2021}
}

@inproceedings{bostrom2020byte,
  title={Byte pair encoding is suboptimal for language model pretraining},
  author={Bostrom, Kaj and Durrett, Greg},
  booktitle={Findings of the Association for Computational Linguistics: EMNLP 2020},
  pages={4617--4624},
  year={2020}
}

@inproceedings{
yuksekgonul2022post,
title={Post-hoc Concept Bottleneck Models},
author={Yuksekgonul, Mert and Wang, Maggie and Zou, James},
booktitle={International Conference on Learning Representations},
year={2023}
}

@inproceedings{oikarinen2022clip,
  title={CLIP-Dissect: Automatic Description of Neuron Representations in Deep Vision Networks},
  author={Oikarinen, Tuomas and Weng, Tsui-Wei},
  booktitle={International Conference on Learning Representations},
  year={2023}
}

\end{document}